\documentclass[letterpaper, 10 pt, conference]{ieeeconf}

\IEEEoverridecommandlockouts                              

\usepackage[utf8]{inputenc}
\usepackage[T1]{fontenc}
\usepackage{csquotes}
\usepackage{hyperref}
\usepackage{graphicx}
\usepackage{multicol, multirow, bigstrut, rotating}
\usepackage{pdfpages}

\usepackage[bibstyle=ieee,
            citestyle=numeric-comp,
            backend=biber,
            doi=false,isbn=false,
            url=true]{biblatex}
\AtEveryBibitem{\clearfield{month}}
\AtEveryCitekey{\clearfield{month}}

\usepackage{todonotes, comment, verbatim}
\newcommand{\etal}{\textit{et al}.}
\newcommand{\eg}{e.g.,}
\newcommand{\ie}{i.e.,}

\usepackage{amsmath, amssymb, bm, units}
\usepackage{mathrsfs}
\usepackage{nicefrac, xfrac, arydshln}
\usepackage{mathtools}

\newcommand{\R}{\mathbb{R}}
\newcommand{\N}{\mathcal{N}}

\newcommand{\mat}[1]{\mathbf{\boldsymbol{#1}}}
\renewcommand{\vec}[1]{\mathbf{\boldsymbol{#1}}}
\newcommand{\smallneg}{\scalebox{0.75}[1.0]{-}}
\newcommand\LG[1]{#1} 
\newcommand\LA[1]{\mathfrak{#1}} 
\newcommand\Lvee[2]{\left[#2\right]^{\vee}_{\LG{#1}}}
\newcommand\Lhat[2]{\left[#2\right]^{\wedge}_{\LG{#1}}}
\newcommand\Lodot[2]{\left[#2\right]^{\odot}_{\LG{#1}}}
\newcommand\LAd[2]{\mathbf{Ad}_{\LG{#1}}\left(#2\right)}
\newcommand\Lad[2]{\mathbf{ad}_{\LG{#1}}\left(#2\right)}
\newcommand\LJac[2]{\Phi_{\LG{#1}}\left(#2\right)}
\newcommand\Lvectran[2]{\exp\left(\Lhat{#1}{#2}\right)}
\newcommand\Ltranvec[2]{\Lvee{#1}{\log\left(#2\right)}}

\newcommand\LveeSmall[2]{[#2]^{\vee}_{\LG{#1}}}
\newcommand\LhatSmall[2]{[#2]^{\wedge}_{\LG{#1}}}

\newcommand\LAdSmall[2]{\mathbf{Ad}_{\LG{#1}}(#2)}
\newcommand\LadSmall[2]{\mathbf{ad}_{\LG{#1}}(#2)}
\newcommand\LJacSmall[2]{\Phi_{\LG{#1}}(#2)}

\newcommand{\diag}{\mathbf{diag}}

\newcommand{\cs}[1]{#1}

\newcommand{\bvraww}[5]{\prescript{\cs{#1}}{#2}{\vec{#3}}^{#4}_{#5}}

\newcommand{\bvUR}[5]{
	\ifthenelse{\equal{#4}{} \OR \equal{#2}{}}{
		\bvraww{#1}{}{#3}{\cs{#2}#4}{#5}
	}{
		( \bvraww{#1}{}{#3}{\cs{#2}}{#5} )^{#4}
	}
}
\newcommand{\bv}[5]{
	\ifthenelse{\equal{#1}{W}}{
		\bvUR{}{#2}{#3}{#4}{#5}
	}{
		\bvUR{#1}{#2}{#3}{#4}{#5}
	}
}
\newcommand{\bvmeas}[5]{
	\bv{#1}{#2}{\breve{#3}}{#4}{#5}
}

\newcommand{\quatUR}[5]{
	\ifthenelse{\equal{#4}{} \OR \equal{#2}{}}{
		\prescript{\cs{#1}}{}{\bm{#3}}^{\cs{#2}#4}_{#5}
	}{
		(\prescript{\cs{#1}}{}{\bm{#3}}^{\cs{#2}}_{#5})^{#4}
	}
}
\newcommand{\quatURR}[5]{
	\ifthenelse{\equal{#1}{W}}{
		\quatUR{}{#2}{#3}{#4}{#5}
	}{
		\quatUR{#1}{#2}{#3}{#4}{#5}
	}
}

\newcommand{\kfsp}[2]{\boldsymbol{\hat{#1}}^{-}_{#2}} 
\newcommand{\kfsm}[2]{\boldsymbol{\hat{#1}}^{+}_{#2}} 
\newcommand{\kfsc}[2]{\boldsymbol{\tilde{#1}}^{+}_{#2}} 
\newcommand{\kfcp}[2]{\mat{#1}^{-}_{#2}} 
\newcommand{\kfcm}[2]{\mat{#1}^{+}_{#2}} 
\newcommand{\dt}{\Delta t}

\title{Estimating Lower Limb Kinematics using \\ a Lie Group Constrained EKF and a Reduced Wearable IMU Count}
\author{Luke Sy$^1$, 
        Nigel H. Lovell$^1$, 
        Stephen J. Redmond$^{1,2}$ 
    \thanks{$^{1}$L. W. Sy, N. H. Lovell, and S. J. Redmond are with the Graduate School of Biomedical Engineering, UNSW Sydney, Australia \{\small l.sy, n.lovell, s.redmond\}@unsw.edu.au }%
    \thanks{$^{2}$S. J. Redmond is with the UCD School of Electrical and Electronic Engineering, University College Dublin, {\small stephen.redmond}@ucd.ie }%
}
\date{4 February 2020}

\begin{document}
	\maketitle
	\thispagestyle{empty}
	\pagestyle{empty}
	
	\begin{abstract}
	    This paper presents an algorithm that makes novel use of a Lie group representation of position and orientation alongside a constrained extended Kalman filter (CEKF) to accurately estimate pelvis, thigh, and shank kinematics during walking using only three wearable inertial sensors. 
		    The algorithm iterates through the prediction update (kinematic equation), measurement update (pelvis height, zero velocity update,  flat-floor assumption, and covariance limiter), and constraint update (formulation of hinged knee joints and ball-and-socket hip joints).
	    The paper also describes a novel Lie group formulation of the assumptions implemented in the said measurement and constraint updates.
	    Evaluation of the algorithm on nine healthy subjects who walked freely within a $4 \times 4 $m$^2$ room shows that the knee and hip joint angle root-mean-square errors (RMSEs) in the sagittal plane for free walking were $10.5 \pm 2.8^\circ$ and $9.7 \pm 3.3^\circ$, respectively, while the correlation coefficients (CCs) were $0.89 \pm 0.06$ and $0.78 \pm 0.09$, respectively.
	    The evaluation demonstrates a promising application of Lie group representation to inertial motion capture under reduced-sensor-count configuration, 
	        improving the estimates (\ie{} joint angle RMSEs and CCs) for dynamic motion, 
	        and enabling better convergence for our non-linear biomechanical constraints.
   		To further improve performance, additional information relating the pelvis and ankle kinematics is needed.
	\end{abstract}

	\section{Introduction}
		Human pose estimation involves tracking the pose (\ie{} position and orientation) of body segments from which joint angles can be calculated.
			It finds application in robotics, virtual reality, animation, and healthcare (\eg{} gait analysis).
		Traditionally, human pose is captured within a laboratory setting using optical motion capture (OMC) systems which can estimate position with up to millimeter accuracy, if well-configured and calibrated.
		    However, recent miniaturization of inertial measurements units (IMUs) has paved the path toward inertial motion capture (IMC) systems suitable for prolonged use outside of the laboratory.
		    
		Commercial IMCs attach one sensor per body segment (OSPS) \cite{Roetenberg2009},
		    which may be considered too cumbersome and expensive for routine daily use by a consumer due to the number of IMUs required.
		    Each IMU typically tracks the orientation of the attached body segment using an orientation estimation algorithm (\eg{} \cite{DelRosario2016a, DelRosario2018}), 
		        which is then connected via linked kinematic chain, usually rooted at the pelvis.
		A reduced-sensor-count (RSC) configuration, where IMUs are placed on a subset of body segments, can improve user comfort while also reducing setup time and system cost.
			However, utilizing fewer sensors inherently reduces the amount of kinematic information available; this information must be inferred by enforcing mechanical joint constraints or making dynamic balance assumptions.
    		Developing a comfortable IMC for routine daily use may facilitate interactive rehabilitation \cite{Llorens2015, Shull2010},
    			and possibly the study of movement disorder progression to enable predictive diagnostics.
		
		RSC performance depends on how the algorithm (i) tracks the body pose, and (ii) infers the kinematic information of these body segments lacking attached sensors.
		    The algorithm may leverage our knowledge of human movement either through data obtained in the past (\ie{} observed correlations between co-movement of different body segments) or by using a simplified model of the human body.
        Data-driven approaches (\eg{} nearest-neighbor search \cite{Tautges2011} and bi-directional recurrent neural network \cite{Huang2018a}) are able to recreate realistic motion suitable for animation-related applications.
            However, these approaches are expected to have a bias toward motions already contained in the database, inherently limiting their use in monitoring pathological gait.
        Model-based approaches reconstruct body motion using kinematic and biomechanical models
            (\eg{} constrained Kalman filter (KF) \cite{sy2020ckf}, extended KF \cite{Lin2012}, particle filter \cite{Meng2012}, and window-based optimization \cite{Marcard2017}).
            Within model-based approaches, using optimization-based estimators can be appealing due to its relative ease to setup and understand.
                However, it can be very inefficient in higher dimensions.
                When estimating the state across time, a recursive estimator can take advantage of the substructure and reduce the state dimension, making the estimator efficient and appropriate for online use \cite{barfoot2017state}.

		Recent work on pose estimation has shown that using a Lie group to represent the states of recursive estimator is a promising approach.
            Such algorithms typically represent the body pose as a chain of linked segments using matrix Lie groups, specifically the special orthogonal group, $SO(n)$, and special Euclidean group, $SE(n)$, where $n=2,3$, are the spatial dimensions of the problem.
		    Traditionally, body poses have been represented using Euler angles or quaternions \cite{Lin2012, Meng2012}.
		Some early work in the field (\cite{Wang2006} and \cite{Barfoot2014}) investigated representations and propagation of pose uncertainty, the former in the context of manipulator kinematics and the latter focused on $\LG{SE(3)}$.
		    This was followed by the formulation of Lie group-based recursive estimators (\eg{} extended KF (EKF) \cite{Bourmaud2014} and unscented KF (UKF) \cite{Brossard2017}).
	    Recently, Lie group based recursive estimators were used to solve the pose estimation problem.
	        Cesic \etal{} estimated pose from marker measurements 
	            and achieved significant improvements compared to an Euler angle representation \cite{Cesic2016};
	            and even supplemented the approach with an observability analysis \cite{Joukov2019}.
		    Joukov \etal{} represented pose using $SO(n)$ with measurements from IMUs under an OSPS configuration.
		        Results also improved, because the Lie group representation is singularity free \cite{Joukov2017}.
    
    	This paper describes a novel human pose estimator that uses a Lie group representation, propagated iteratively using a CEKF to estimate lower body kinematics for an RSC configuration of IMUs.
    	    It builds on prior work \cite{sy2020ckf} but instead represents the state variables as Lie groups, specifically $\LG{SE(3)}$, to track both position and orientation (\cite{sy2020ckf} only tracks position).
    	    Furthermore, this paper describes a novel Lie group formulation for assumptions specific to pose estimation, such as zero velocity update, and biomechanical constraints (\eg{} constant thigh length and a hinged knee joint).
    	    Note that this algorithm is different from \cite{Joukov2017} in that the state (\ie{} body pose) was represented as $\LG{SE(3)}$ instead of $SO(n)$.
    	        This representation allows for tracking of the global position of the body, incorporating IMU measurements in the prediction step, and a simpler  implementation of measurement assumptions at the cost of requiring an additional constraint step.
	    The design was motivated by the need for a better state variable representation which would potentially better model the biomechanical system to infer the missing kinematic information from uninstrumented body segments. 
	        Advancing such algorithms can lead to the development of a gait assessment tool using as few sensors as possible, ergonomically-placed for comfort, to facilitate long-term monitoring of lower body movement.
            
	\section{Algorithm description}
\begin{figure}[htbp]
	        \centering
	        \includegraphics[width=0.35\textwidth]{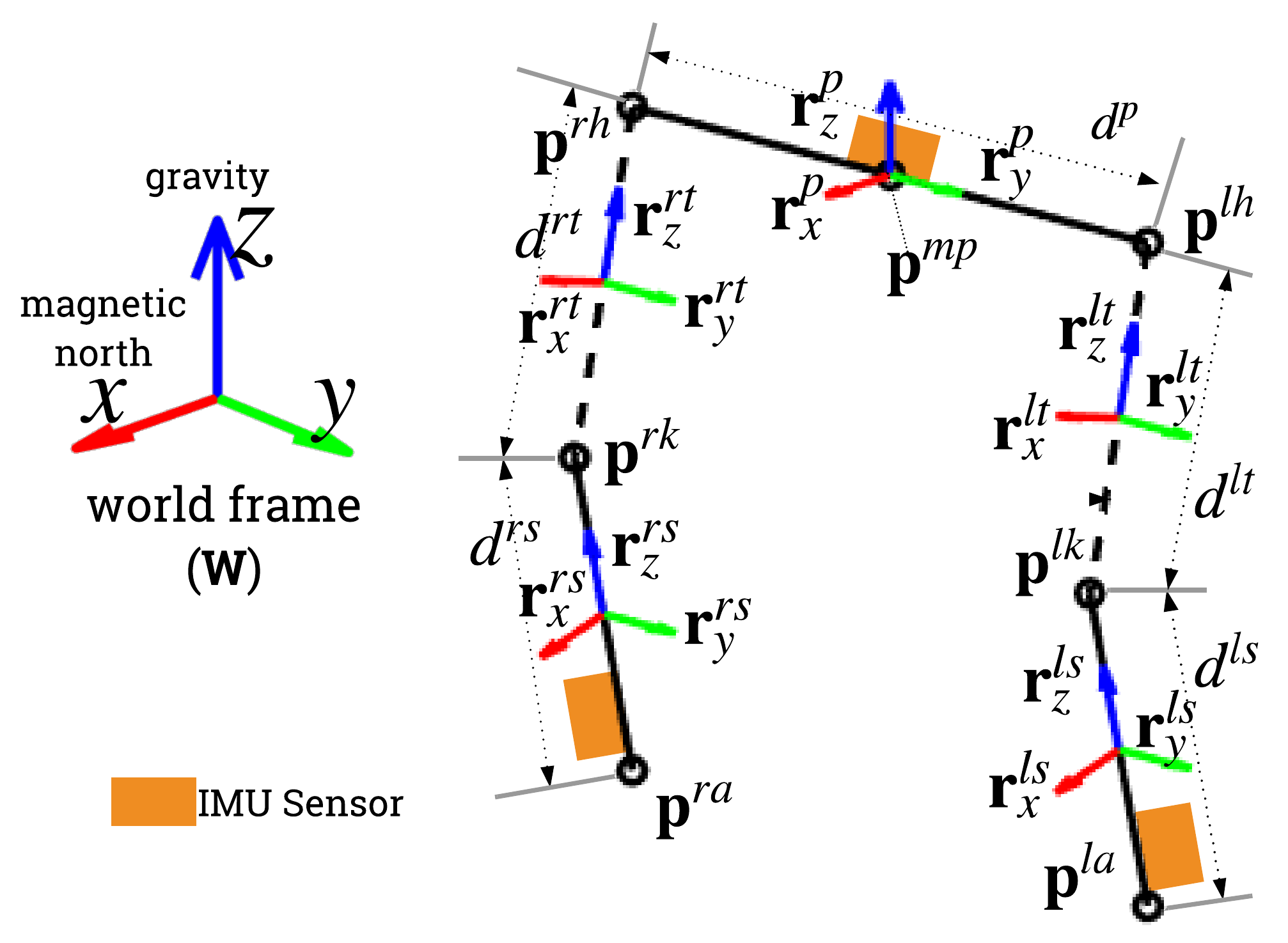}
	        \caption{Physical model of the lower body used by the algorithm. The circles denote joint positions, the solid lines denote instrumented body segments, whilst the dashed lines denote segments without IMUs attached (\ie{} thighs).}
	        \label{fig:body-skeleton}
	    \end{figure}

\begin{figure}[htbp]
	        \centering
	        \includegraphics[width=0.45\textwidth]{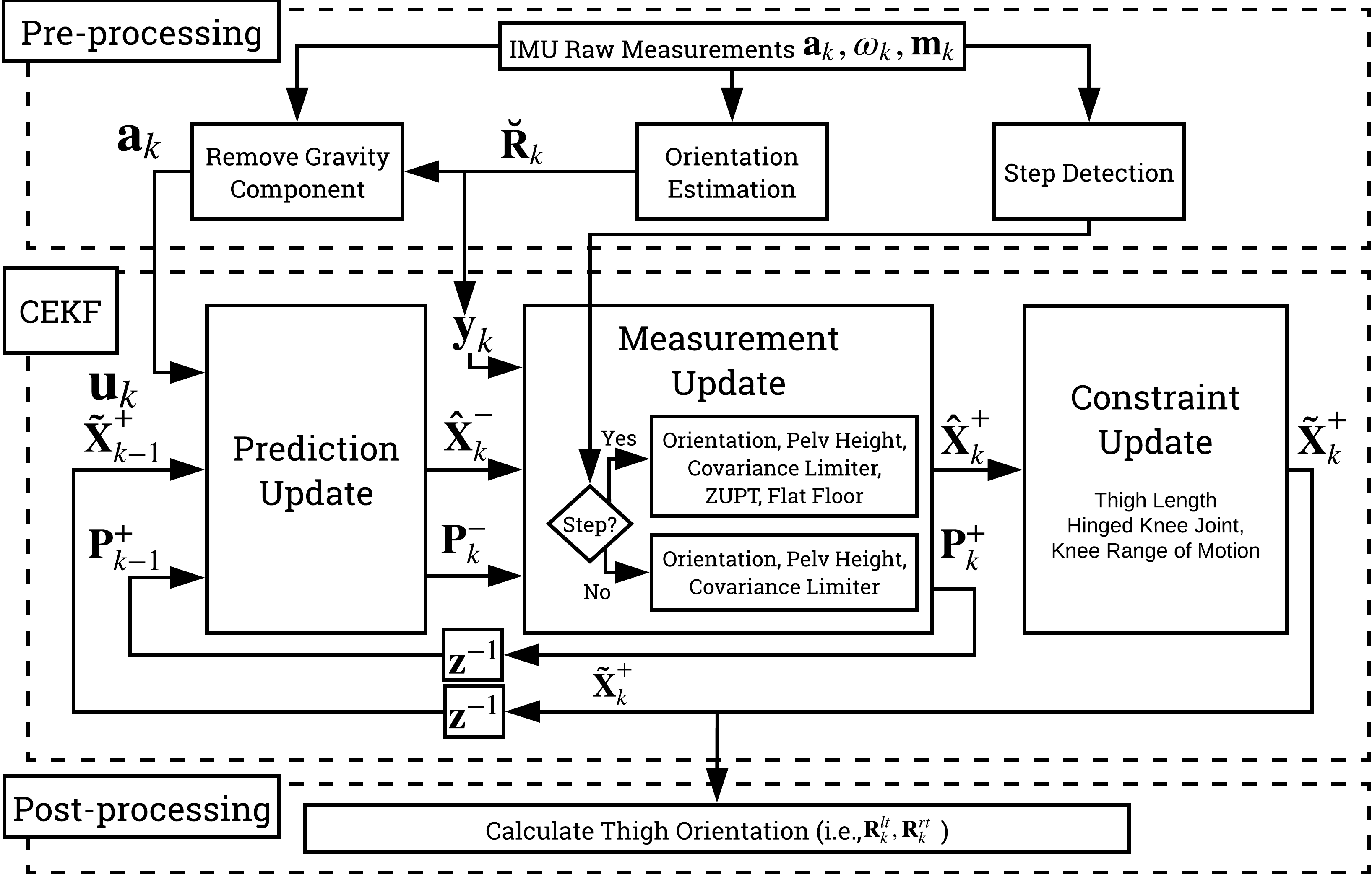}
	        \caption{Algorithm overview which consists of pre-processing, CEKF, and post-processing. 
	            Pre-processing calculates the body segment orientation, inertial body acceleration, and step detection from raw acceleration, $\vec{a}_{k}$, angular velocity, $\vec{\omega}_{k}$, and magnetic north heading, $\vec{h}_{k}$, measured by the IMU. 
	            The CEKF-based state estimation consists of a prediction (kinematic equation), measurement (orientation, pelvis height, covariance limiter, intermittent zero-velocity update, and flat-floor assumption), and constraint update (thigh length, hinge knee joint, and knee range of motion). 
	            Post-processing calculates the left and right thigh orientations, $\mat{R}_{lt}$ and $\mat{R}_{rt}$.}
	        \label{fig:algo-overview}
	         
	    \end{figure}

	    The proposed algorithm, \emph{LGKF-3IMU}, uses a similar model and assumptions to our prior work in \cite{sy2020ckf}, denoted as \emph{CKF-3IMU}, albeit expressed in Lie group representation, 
	        to estimate the orientation of the pelvis, thighs, and shanks with respect the world frame, $\cs{W}$, using only three IMUs attached at the sacrum and shanks, just above the ankles (Fig. \ref{fig:body-skeleton}).
	    Using a Lie group representation enables the tracking of not just position but also of orientation singularity free (note that \emph{CKF-3IMU} only tracked position and assumed orientation as perfect),
	        whilst improving performance for dynamic movements and utilizing fewer assumptions.
        Fig. \ref{fig:algo-overview} shows an overview of the proposed algorithm.
        \emph{LGKF-3IMU} predicts the shank and pelvis positions through double integration of their linear 3D acceleration as measured by the attached IMUs (after a pre-processing step that resolves these accelerations in the world frame).
            Orientation is obtained from a third party orientation estimation algorithm.
            To mitigate positional drift due to sensor noise that accumulates in the double integration of acceleration, the following assumptions are enforced:
                (1) the ankle 3D velocity and height above the floor are zeroed whenever a footstep is detected;
                (2) the pelvis Z position is approximated as the length of the unbent leg(s) above the floor.
            Furthermore, to control the otherwise ever-growing error covariance for the pelvis and ankle positions, a pseudo-measurement equal to the current pose state estimate with a fixed covariance is made. 
            Lastly, biomechanical constraints enforce constant body segment length; ball-and-sockets hip joints; and a hinge knee joint (one degree of freedom (DOF)) with limited range of motion (ROM).
        The pre- and post-processing parts remains exactly the same as the \emph{CKF-3IMU} algorithm.
        
	    
    \subsection{Lie group and Lie algebra}
    	The matrix Lie group $\LG{G}$ is a group of $n \times n$ matrices that is also a smooth manifold (\eg{} $\LG{SE(3)}$).
        	Group composition and inversion (\ie{} matrix multiplication and inversion) are smooth operations.
    	Lie algebra $\LA{g}$ represents a tangent space of a group at the identity element \cite{selig2004lie}.
    	    The elegance of Lie theory lies in it being able to represent curved objects using a vector space (\eg{} Lie group $\LG{G}$ represented by $\LA{g}$) \cite{stillwell2008naive}.
    	        	
\begin{figure}
            \centering
            \resizebox{0.7\linewidth}{!}{
			\begin{tikzpicture}
				\begin{scope}
				\clip (-2.5,0) rectangle (2.5,2.5);
				\draw (0,0) circle (2.5) node[above left] (A) {Lie group $\LG{G}$};
				\end{scope}
				\draw[<->] (1.77-1.5,1.77+1.5) -- (1.77+1.5,1.77-1.5)
				node[pos=0.5, label={above right:{Lie algebra $\LA{g}$}}] (B) {};
				\draw[<->] (3.5,2.5) -- (5.5,2.5)
				node[pos=0.7, above] (C) {$\R^n$};
				\draw[->] (A) .. controls +(up:1cm) and +(left:1cm) .. (B)
				node[pos=0.5, above]{$\log{}_{\LG{G}}$};
				\draw[->] (B) .. controls +(down:1cm) and +(right:2cm) .. (A)
				node[pos=0.5, below right]{$\exp{}_{\LG{G}}$};
				\draw[->] (B) .. controls +(up:2cm) and +(up:1cm) .. (C)
				node[pos=0.5, below]{$\Lvee{G}{\:\:}$};
				\draw[->] (C) .. controls +(down:2cm) and +(right:1cm) .. (B)
				node[pos=0.5, below]{$\Lhat{G}{\:\:}$};
			\end{tikzpicture}}
            \caption{Mapping between Lie group $\LG{G}$, Lie algebra $\LA{g}$, and a $n$-dimensional vector space.}
            \label{fig:lie-group-algebra-overview}
        \end{figure}
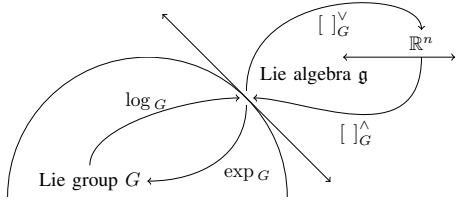

    	The matrix exponential $\exp{}_{\LG{G}}: \LA{g} \tiny{\to} \LG{G}$ and matrix logarithm $\log{}_{\LG{G}}: \LG{G} \tiny{\to} \LA{g}$ establish a local diffeomorphism between the Lie group $\LG{G}$ and its Lie algebra $\LA{g}$.
    	   The Lie algebra $\LA{g}$ is a $n \times n$ matrix that can be represented compactly with an $n$ dimensional vector space. A linear isomorphism between $\LA{g}$ and $\R^n$ is given by
                $\Lvee{G}{\:\:}: \LA{g} \tiny{\to} \R^n$ and
                $\Lhat{G}{\:\:}: \R^n \tiny{\to} \LA{g}$.
            An illustration of the said mappings are given in Fig. \ref{fig:lie-group-algebra-overview}.
        Furthermore, the adjoint operators of a Lie group, denoted as $\LAdSmall{G}{X}$, and Lie algebra, denoted as $\LadSmall{G}{X}$ will be used in later sections.
            For a more detailed introduction to Lie groups refer to \cite{barfoot2017state, stillwell2008naive, Chirikjian2012Book2}.
	        
	\subsection{System, measurement, and constraint models}
  		The system and measurement models are presented below
    	    \begin{gather}
    	        \vec{X}_{k} = f(\vec{X}_{k\smallneg1}, \vec{n}_{k\smallneg1}) = \vec{X}_{k\smallneg1} \exp_{G} ( \Lhat{G}{\Omega (\vec{X}_{k\smallneg1} ) \tiny{+} \vec{n}_{k\smallneg1}} ) \label{eq:pred-update} \\
    	        \vec{Z}_{k} = h (\vec{X}_{k}) \exp_{G}\left(\Lhat{G}{\vec{m_{k}}}\right)  ,\:\:
    	        \vec{D}_{k} = c (\vec{X}_{k}) \label{eq:meas-cstr-update}
            \end{gather}
        where $k$ is the time step; 
        	$\vec{X}_{k} \in \LG{G}$ is the system state, an element of state Lie group $\LG{G}$;
        	$\Omega \left(\vec{X}_{k}\right) : \LG{G} \tiny{\to} \R^p$ is a non-linear function;
        	$\vec{n}_{k}$ is a zero-mean process noise vector with covariance matrix $\mat{Q}_{k}$ (\ie{} $\vec{n}_k \sim \N_{\R^p}(\vec{0}_{p \times 1}, \mat{Q}_{k})$);
        	$\vec{Z}_{k} \in \LG{G_1}$ is the system measurement, an element of measurement Lie group $\LG{G_1}$;
        	$ h\left(\vec{X}_{k}\right): \LG{G} \tiny{\to} \LG{G_1}$ is the measurement function;
        	$\vec{m}_{k}$ is a zero-mean measurement noise vector with covariance matrix $\mat{R}_{k}$ (\ie{} $\vec{m}_k \sim \N_{\R^q}(\vec{0}_{q \times 1}, \mat{R}_{k})$);
        	$\mat{D}_{k} \in \LG{G_2}$ is the constraint state,
        	 an element of constraint Lie group $\LG{G_2}$;
        	$ c\left(\vec{X}_{k}\right): \LG{G} \tiny{\to} \LG{G_2}$ is the equality constraint function the state $\vec{X}_{k}$ must satisfy.
        Similar to \cite{Bourmaud2013, Cesic2016}, the state distribution of $\vec{X}_{k}$ is assumed to be a concentrated Gaussian distribution on Lie groups (\ie{} $\vec{X}_{k} = \vec{\mu}_k \exp_{\LG{G}} \Lhat{G}{\vec{\epsilon}}$ where $\vec{\mu}_k$ is the mean of $\vec{X}_{k}$ and Lie algebra error $\vec{\epsilon} \sim \N_{\R^p}(\vec{0}_{p \times 1}, \mat{P})$) \cite{Wang2006}.
        The Lie group state variables $\vec{X}_{k}$ model the position, orientation, and velocity of the three instrumented body segments (\ie{} pelvis and shanks) as 
        $\vec{X}_k = \diag \left( 
        		\bv{W}{p}{T}{}{}, \bv{W}{ls}{T}{}{}, \bv{W}{rs}{T}{}{},
        		\bv{W}{p}{\mathring{v}}{}{}, \bv{W}{ls}{\mathring{v}}{}{}, \bv{W}{rs}{\mathring{v}}{}{}
       		\right)$ $\in$ $ \LG{G} = \LG{SE(3)}^3 \times \R^{9}$
        	where $\bv{A}{B}{T}{}{} \in \LG{SE(3)}$ denotes the pose of body segment $B$ relative to frame $A$,
        	    and $\mathring{\vec{v}^x}=\begin{bsmallmatrix} I_{3 \times 3} & \vec{v}^x \\ 0_{1 \times 3} & 1 \end{bsmallmatrix}$ is the trivial mapping of a 3D vector to an element in $SE(3)$.
        	    If frame $\cs{A}$ is not specified, assume reference to the world frame, $\cs{W}$.
        	$\Lvee{}{\:}$, $\Lvectran{G}{\:}$, $\Ltranvec{G}{\:}$, and $\LAd{}{\vec{X}_k}$ are constructed similarly.
        	See \cite{barfoot2017state} for $\LG{SE(3)}$ operator definitions.

    \subsection{Lie group constrained EKF (LG-CEKF)}
        The \textit{a priori} (predicted), \textit{a posteriori} (updated using measurements), and constrained state (satisfying the state constraint equation, \ie{} biomechanical constraints) for time step $k$ are denoted by $\kfsp{\mu}{k}$, $\kfsm{\mu}{k}$, and $\kfsc{\mu}{k}$, respectively. 
        The KF state error \textit{a priori} and \textit{a posteriori} covariance matrices are denoted as $\kfcp{P}{k}$ and $\kfcm{P}{k}$, respectively.
		The KF is based on the Lie group EKF, as defined in \cite{Bourmaud2013}.
		
    \subsubsection{Prediction step} \label{sec:pred-update}
        estimates the \textit{a priori} state $\kfsp{\mu}{k}$ at the next time step and may not necessarily respect the kinematic constraints of the body, so joints may become dislocated after this prediction step.
        The mean propagation of the three instrumented body segments is governed by Eq. \eqref{eq:lgkf-predmu} 
            where $\kfsc{\Omega}{k} = \Omega (\kfsc{\mu}{k})$ and $\Omega (\mat{X}_{k})$ is the motion model for the three instrumented body segments.
            For the sake of brevity, only the motion model of the position, orientation, and velocity for body segment $b$ is shown (Eqs. \eqref{eq:lgkf-omegab}).
                The measured acceleration and orientation of segment $B$ are denoted as $\bvmeas{W}{B}{a}{}{k}$ and $\bvmeas{W}{B}{R}{}{k}$.
            The process noise for body segment $b$ is shown in Eq. \eqref{eq:lgkf-omeganoise} where $\bv{}{b}{\sigma}{}{acc}$ and $\bv{}{b}{\sigma}{}{qori}$ denote the noise variances of the measured acceleration and orientation.
            Note that one may use the measured angular velocity to predict orientation.
                However, we chose setting angular velocity to zero to simplify computations related to position, knowing that the orientation will be updated in the measurement step using measurements from a third party orientation estimation algorithm, accounting for angular velocity.
	    \begin{gather}
        	\kfsp{\mu}{k+1} = \kfsc{\mu}{k} \exp_{G} ( 
        							\LhatSmall{G}{\kfsc{\Omega}{k}} 
       							) \label{eq:lgkf-predmu} \\
        	\Omega^{b} \left( \mat{X}_{k} \right) = [
		        	(
		        	\dt \bv{W}{b}{v}{}{k} +
		        	\tfrac{\dt^2}{2} \bvmeas{W}{b}{a}{}{k} 
		        	)^T \bvmeas{W}{b}{R}{}{k} \:\:
		        	\bv{}{}{0}{}{1 \times 3} \:\:
		        	\dt \bvmeas{W}{b}{a}{T}{k} 
	        ]^T \label{eq:lgkf-omegab}\\
	       	\vec{n}^{b} = [
	       	    \tfrac{\dt^2}{2} \bv{}{b}{\sigma}{T}{acc} \quad
			    \bv{}{b}{\sigma}{T}{qori} \quad
			    \dt \bv{}{b}{\sigma}{T}{acc}
	       	]^T \label{eq:lgkf-omeganoise}
        \end{gather}
        
        The state error covariance matrix propagation is governed by Eq. \eqref{eq:lgkf-predP} 
            where $\mathcal{F}_{k}$ represents the matrix Lie group equivalent to the Jacobian of $f(\vec{X}_{k\smallneg1}, \vec{n}_{k\smallneg1})$,
            $\mathscr{C}_{k}$ represents the linearization of the motion model,
            $\mat{Q}_{k}$ is constructed from with diagonal values from $\vec{n}^{b}$,
            and $\bv{}{}{\mu}{\epsilon}{k} = \bv{}{}{\mu}{}{k} \exp_{G}(\Lhat{G}{\vec{\epsilon}})$ represents the state  with infinitesimal perturbation $\vec{\epsilon}$.
        Refer to the supplementary material \cite{sy2020lgcekfsupp} for the explicit definition of the motion model, $\Omega_{k} \left( \mat{X}_{k} \right)$, and $\mathscr{C}_{k}$.
        \begin{gather}
    	    \kfcp{P}{k+1} = \mathcal{F}_{k} \kfcm{P}{k} \mathcal{F}_{k}^T + \LJacSmall{G}{\hat{\Omega}_k} \mat{Q}_{k} \LJacSmall{G}{\hat{\Omega}_k}^T \label{eq:lgkf-predP} \\
    	    \mathcal{F}_{k} = \LAdSmall{G}{\exp_{G} ( -\LhatSmall{G}{\hat{\Omega}_k} )} + \LJacSmall{G}{\hat{\Omega}_k} \mathscr{C}_k \\
    	    \mathscr{C}_{k} = \tfrac{\partial}{\partial \vec{\epsilon}} 
    	    \Omega\left( \vec{\mu}_{k}^\epsilon \right) |_{\vec{\epsilon} = 0}, \\
    	    \LJac{G}{\vec{X}_k} = \textstyle\sum_{i=0}^{\infty} \tfrac{(-1)^i}{(i+1)!} \Lad{G}{\vec{X}_k}^i
	    \end{gather}

    \subsubsection{Measurement update} \label{sec:meas-update}
       estimates the state at the next time step by: 
        	(i) updating the orientation state using new orientation measurements of body segments;
        	(ii) encouraging pelvis Z position to be close to initial standing height $z_{p}$, and by;
        	(iii) encouraging ankle velocity to approach zero, and the ankle Z position to be close to the floor level, $z_{f}$.
        The \textit{a posteriori} state $\kfsm{\mu}{k}$ is calculated following the Lie EKF equations below.
            $\mathcal{H}_{k}$ can be seen as the matrix Lie group equivalent to the Jacobian of $h(\mat{X}_k)$; 
                and is defined as the concatenation of $\mathcal{H}_{ori}$ and $\mathcal{H}_{mp}$.  $\mathcal{H}_{ls}$ and/or $\mathcal{H}_{rs}$ are also concatenated to $\mathcal{H}_{k}$ when the left and/or right foot contact is detected (See \cite[Eq. (9)]{sy2020ckf}).
                Each component matrix will be described later.
        	$\mat{Z}_k$, $h\left( \mat{X}_k \right)$, and $\mat{R}_k$ are constructed similarly to $\mathcal{H}_{k}$ but combined using $\diag$ instead of concatenation 
        	(\eg{} $\mat{R}_{k} = \diag(\vec{\sigma}_{ori}, \vec{\sigma}_{mp})$)
        \begin{gather}
	        \mat{K}_{k} = \kfcp{P}{k} \mathcal{H}_{k}^T ( \mathcal{H}_{k} \kfcp{P}{k} \mathcal{H}_{k}^T + \mat{R}_{k} )^{-1} \\
	        \vec{\nu}_{k} = \mat{K}_{k} ( \LveeSmall{G_1}{\log_{G_1} \left( h(\kfsp{\mu}{k})^{-1} \mat{Z}_k \right) } ) \\
	        \kfsm{\mu}{k} = \kfsp{\mu}{k} \exp_{G} ( \Lhat{G}{\vec{\nu}_k} ) \\
	        \mathcal{H}_{k} = \tfrac{\partial}{\partial \vec{\epsilon}} \LveeSmall{G_1}{
	        	\log_{G_1} \big( 
	        	h( \kfsp{\mu}{k} )^{-1} 
	        	h( \vec{\mu}_k^\epsilon ) \big) } |_{\vec{\epsilon}=0} \label{eq:dHdef}
        \end{gather}
        
    	The measurement functions of the (i) orientation update, (ii) pelvis height assumption, and (iii) ankle velocity and flat floor assumptions are defined by Eqs. \eqref{eq:h-ori-k}-\eqref{eq:hZ-lstep-k} with measurement noise variances $\vec{\sigma}_{ori}^2$ ($9 \times 1$ vector), $\vec{\sigma}_{mp}^2$ ($1 \times 1$ vector), and $\vec{\sigma}_{ls}^2$  ($4 \times 1$ vector), respectively.
        	$\mat{I}_{i \times j}$ and $\mat{0}_{i \times j}$ denote $i \times j$ identity and zero matrices;
        	$\vec{i}_x$, $\vec{i}_y$, $\vec{i}_z$, and $\vec{i}_0$ denote $4 \times 1$ vectors whose 1$^{st}$ to 4$^{th}$ row, respectively, are $1$, while the rest are $0$;
        	and the $^\odot$ operator is as defined in \cite[][Eq. (72)]{barfoot2017state}.
            $\mathcal{H}_{ori}$, $\mathcal{H}_{mp}$, and $\mathcal{H}_{ls}$ (Eqs. \eqref{eq:H-ori-k}-\eqref{eq:H-lstep-k}) are calculated by applying Eq. \eqref{eq:dHdef} to their corresponding measurement function, 
                followed by tedious algebraic manipulation and first order linearization (\ie{} $\exp(\Lhat{}{\epsilon}) \approx \mat{I} + \Lhat{}{\epsilon}$).
                See details in the supplementary material \cite{sy2020lgcekfsupp}.
        \begin{gather}
        	h_{ori} \left( \mat{X}_{k} \right) = \diag(\bv{W}{p}{R}{}{k}, \bv{W}{ls}{R}{}{k}, \bv{W}{rs}{R}{}{k}) \label{eq:h-ori-k}\\
        	\mat{Z}_{ori} = \diag(\bvmeas{W}{p}{R}{}{k}, \bvmeas{W}{ls}{R}{}{k}, \bvmeas{W}{rs}{R}{}{k}) \label{eq:Z-ori-k} \\
			h_{mp} \left( \mat{X}_{k} \right) = \vec{i}_z^T \bv{W}{p}{T}{}{} \vec{i}_0 ,\quad
				\mat{Z}_{mp} = z_{p} \label{eq:hZ-mp-k} \\
			h_{ls} \left( \mat{X}_{k} \right) = \begin{bmatrix} 
				\bv{W}{ls}{v}{}{} \\
				\vec{i}_z^T \bv{W}{ls}{T}{}{} \vec{i}_0
			\end{bmatrix} ,\quad
			\mat{Z}_{ls} = \begin{bmatrix} \mat{0}_{3 \times 1} \\ z_{f} \end{bmatrix} \label{eq:hZ-lstep-k} \\
        	\mathcal{H}_{ori} = \left[ \begin{array}{ccc:c}
        		\mat{0}_{3 \times 3} \: \mat{I}_{3 \times 3} & & & \\
        		& \mat{0}_{3 \times 3} \: \mat{I}_{3 \times 3} & & \mat{0}_{9 \times 9} \\
	        	& & \mat{0}_{3 \times 3} \: \mat{I}_{3 \times 3} & \\
	        	\end{array} \right] \label{eq:H-ori-k} \\
            \mathcal{H}_{mp} = \left[ \begin{array}{ccc:c}
				\vec{i}_z^T \bv{W}{p}{\bar{T}}{}{} \Lodot{}{\vec{i}_0} & \mat{0}_{1 \times 6} & \mat{0}_{1 \times 6} & \mat{0}_{1 \times 9} \\
			\end{array} \right] \label{eq:H-mp-k} \\
            \mathcal{H}_{ls} = \left[ \begin{array}{c:c:c:c:c}
				\multirow{2}{*}{$\dots$} & \small\text{pos. ori. col.} & \multirow{2}{*}{$\dots$} & 
				\underbrace{\mat{I}_{3 \times 3}} & \multirow{2}{*}{$\dots$} \\
				& \overbrace{\vec{i}_z^T \bv{W}{ls}{\bar{T}}{}{} \Lodot{}{\vec{i}_0}} & & \small\text{vel. col.} & \\
			\end{array} \right] \label{eq:H-lstep-k}
		\end{gather}     
        Lastly, the covariance limiter prevents the covariance from growing indefinitely and from becoming badly conditioned, as will happen naturally when tracking the global position of the pelvis and ankles without any global position reference.
            At this step, a pseudo-measurement equal to the current state $\kfsm{\mu}{k}$ is used (implemented by $\mathcal{H}_{lim} = \begin{bmatrix} \mat{I}_{18 \times 18} & \mat{0}_{18 \times 9} \end{bmatrix}$) with some measurement noise of variance $\bv{}{}{\sigma}{}{lim}$ ($9 \times 1$ vector).
            The covariance $\kfcm{P}{k}$ is then calculated through Eqs. \eqref{eq:Hkprime}-\eqref{eq:Ptildekprime}.
        \begin{gather}
            \mathcal{H}_{k}' = \begin{bmatrix} \mathcal{H}_{k}^T & \mathcal{H}_{lim}^T \end{bmatrix}^T, \quad
            \mat{R}_{k}' = \diag([ \bv{}{}{\sigma}{}{k} \: \bv{}{}{\sigma}{}{lim} ]) \label{eq:Hkprime} \\
            \mat{K}_{k}' = \kfcp{P}{k} \mathcal{H}_{k}'^T \left( \mathcal{H}_{k}' \kfcp{P}{k} \mathcal{H}_{k}'^T + \mat{R}' \right)^{-1} \label{eq:Kkprime} \\
            \kfcm{P}{k} = \LJac{G}{\vec{\nu}_k} \left(\mat{I} - \mat{K}_{k}' \mathcal{H}_{k}' \right) \kfcp{P}{k} \LJac{G}{\vec{\nu}_k}^T \label{eq:Ptildekprime}
        \end{gather}
        
    \subsubsection{Satisfying biomechanical constraints} \label{sec:const-nonlin-update}
        After the prediction and measurement updates, above, the body joints may have become dislocated, or joint angles extend beyond their allowed range. This update corrects the kinematic state estimates to satisfy the biomechanical constraints of the human body by projecting the current \textit{a posteriori} state $\kfsm{\mu}{k}$ estimate onto the constraint surface, guided by our uncertainty in each state variable, encoded by $\kfcm{P}{k}$.   
        	The constraint equations enforce the following biomechanical limitations: 
        		(i) the length of estimated thigh vectors ($||\bv{W}{lt}{\tau}{}{}||$ and $||\bv{W}{rt}{\tau}{}{}||$) equal the thigh lengths $d^{\cs{lt}}$ and $d^{\cs{rt}}$;
        		(ii) both knees act as hinge joints (formulation similar to \cite[Sec. 2.3 Eqs. (4)]{Meng2012}); 
        		and (iii) the knee joint angle is confined to realistic ROM.
        The constrained state $\kfsc{\mu}{k}$ can be calculated using the equations below, similar to the measurement update of \cite{Bourmaud2013} with zero noise 	
    		where $\mathcal{C}_{k} = \begin{bmatrix} \mathcal{C}_{L, k}^T & \mathcal{C}_{R, k}^T \end{bmatrix}^T$. 
    			$\mathcal{C}_{L, k}$ is the concatenation of $\mathcal{C}_{ltl,k}$, $\mathcal{C}_{lkh,k}$, and $\mathcal{C}_{lkr,k}$; the last matrix is not concatenated when the knee angle, $\alpha_{lk}$, is bounded (\ie{} $\alpha_{lk,min} \leq \alpha_{lk} \leq \alpha_{lk,max}$).
    			Each component matrix will be described later.
    			$\mathcal{C}_{R, k}$ can be derived similarly,
    			while $\mat{D}_k$ and $c \left( \mat{X}_k \right)$ are constructed similarly to $\mat{Z}_k$.
		    \begin{gather}
			    \mat{K}_{k} = \kfcm{P}{k} \mathcal{C}_{k}^T ( \mathcal{C}_{k} \kfcm{P}{k} \mathcal{C}_{k}^T )^{-1} \\
			    \vec{\nu}_{k} = \mat{K}_{k} ( \LveeSmall{G_2}{\log_{G_2} ( c(\kfsm{\mu}{k})^{-1} \mat{D}_k ) }) \\
			    \kfsc{\mu}{k} = \kfsm{\mu}{k} \exp_{G} \left( \Lhat{G}{\vec{\nu}_k} \right) \\
			    \mathcal{C}_{k} = \tfrac{\partial}{\partial \vec{\epsilon}} \LveeSmall{G_2}{
			    	\log_{G_2} \big( 
			    	c\left( \kfsm{\mu}{k} \right)^{-1} 
			    	c\left( \vec{\mu}_k^\epsilon \right) \big) 
			    	} |_{\vec{\epsilon}=0} \label{eq:dCdef}
		    \end{gather}
			
		The constraint functions are similar to \cite[Sec. II-E.3]{sy2020ckf} but expressed under  $\LG{SE(3)}$ state variables.
    		Firstly, the thigh length constraint is shown in Eq. \eqref{eq:c-lthigh} where $\bv{W}{lt}{\tau}{}{z}(\kfsc{\mu}{k})$ denotes the thigh vector.
    		Secondly, the hinge knee joint constraint is defined by Eq. \eqref{eq:c-lkhinge}.
	        Thirdly, the knee ROM constraint is defined by Eq. \eqref{eq:c-lkrom-int3} and is only enforced if the knee angle, $\alpha_{lk}$, is outside the allowed ROM.
	        The bounded knee angle, $\alpha_{lk}'$, is calculated by Eqs. \eqref{eq:c-lkrom-cstr} and \eqref{eq:c-lkrom-base}.
		Lastly. $\mathcal{C}_{ltl,k}$, $\mathcal{C}_{lkh,k}$, and $\mathcal{C}_{lkr,k}$ are calculated by applying Eq. \eqref{eq:dCdef} to their corresponding constraint functions, similar to $\mathcal{H}_{mp}$.
		    Refer to the supplementary material for full derivation \cite{sy2020lgcekfsupp}.
        \begin{gather}
	        \bv{p}{lh}{p}{}{} = \begin{bmatrix}
		        0 & \tfrac{d^{\cs{p}}}{2} & 0 & 1
		        \end{bmatrix}^T ,\quad
	        \bv{ls}{lk}{p}{}{} = \begin{bmatrix}
		        0 & 0 & d^{\cs{ls}} & 1
		        \end{bmatrix}^T \\
	        \bv{W}{lt}{\tau}{}{z}(\kfsc{\mu}{k}) = 
	        	\overbrace{\begin{bmatrix}
	        		\mat{I}_{3 \times 3} & \mat{0}_{3 \times 1}
	        		\end{bmatrix}}^{\mat{E}} \big(
		        \overbrace{\bv{W}{p}{T}{}{} \bv{p}{lh}{p}{}{} }^{\text{hip joint pos.}} - 
		        \overbrace{\bv{W}{ls}{T}{}{} \bv{ls}{lk}{p}{}{} }^{\text{knee joint pos.}} \big) \label{eq:thigh-vect} \\
	        c_{ltl}(\kfsc{\mu}{k}) 
	        	= \bv{W}{lt}{\tau}{}{z}(\kfsc{\mu}{k})^T 			
	        \bv{W}{lt}{\tau}{}{z}(\kfsc{\mu}{k}) - (d^{\cs{lt}})^2 
	        	= 0 = \mat{D}_{ltl} \label{eq:c-lthigh} \\
            c_{lkh}(\kfsc{\mu}{k}) = \bv{W}{ls}{r}{T}{y} \bv{W}{lt}{\tau}{}{z}
            	= 0 = \mat{D}_{lkh} \label{eq:c-lkhinge} \\
 			\alpha_{lk}' = \mathbf{min}(\alpha_{lk,max}, \mathbf{max}(\alpha_{lk,min}, \alpha_{lk})) \label{eq:c-lkrom-cstr} \\
			\alpha_{lk} = \tan^{-1} \left( 
			\tfrac{ -\bv{W}{ls}{r}{T}{z} \bv{W}{lt}{r}{}{z} }{ -\bv{W}{ls}{r}{T}{x} \bv{W}{lt}{r}{}{z} } 
			\right) + \tfrac{\pi}{2} \label{eq:c-lkrom-base} \\
			\begin{split}
				c_{lkr}(\kfsc{\mu}{k}) &= 
					(\bv{W}{ls}{r}{T}{z} \tiny\cos(\alpha_{lk}' \text{--} \tfrac{\pi}{2}) \text{--} 
				     \bv{W}{ls}{r}{T}{x} \tiny\sin(\alpha_{lk}' \text{--} \tfrac{\pi}{2}))
					\bv{W}{lt}{r}{}{z} \\
				&= 0 = \mat{D}_{lkr}
			\end{split} \label{eq:c-lkrom-int3}
		\end{gather}

	\section{Experiment}
    	The dataset from \cite{sy2020ckf} was used to evaluate \emph{LGKF-3IMU}.
            It involved movements listed in Table \ref{tab:movement-type-desc} from nine healthy subjects ($7$ men and $2$ women, weight $63.0 \pm 6.8$ kg, height $1.70 \pm 0.06$ m, age $24.6 \pm 3.9$ years old), with no known gait abnormalities.
                Raw data were captured using a commercial IMC (\ie{} Xsens Awinda) compared against a benchmark OMC (\ie{} Vicon) within an \texttildelow$4 \times 4$ m$^2$ capture area.
        \begin{table}[htbp]
            \centering
            \caption{Types of movements done in the validation experiment}
            \label{tab:movement-type-desc}
            \begin{tabular}{lllc}
                \hline \hline
                 Movement & Description & Duration & Group \\ \hline 
                 Walk & Walk straight and return & $\sim 30$ s & F\\
                 Figure-of-eight & Walk along figure-of-eight path & $\sim 60$ s & F \\
                 Zig-zag & Walk along zig-zag path & $\sim 60$ s & F \\
                 5-minute walk & Unscripted walk and stand & $\sim 300$ s & F \\
                 Speedskater & Speedskater on the spot & $\sim 30$ s & D \\
                 Jog & Jog straight and return & $\sim 30$ s & D \\
                 Jumping jacks & Jumping jacks on the spot & $\sim 30$ s & D \\
                 High knee jog & High knee jog on the spot & $\sim 30$ s & D \\
                 \hline \hline
            \end{tabular}
            
            F denotes free walk, D denotes dynamic
        \end{table}
        
        Unless stated, calibration and system parameters similar to \cite{sy2020ckf} were assumed.
           	The algorithm and calculations were implemented using Matlab 2018b.
            The initial position, orientation, and velocity ($\kfsc{\mu}{0}$) were obtained from the Vicon benchmark system.
            $\kfcm{P}{0}$ was set to $0.5 \mat{I}_{27 \times 27}$.
	            The variance parameters used to generate the process and measurement error covariance matrix $\mat{Q}$ and $\mat{R}$ are shown in Table \ref{tab:var-param-kfnoise}.
            \begin{table}[htbp]
                \centering
                \caption{Variance parameters for generating the process and measurement error covariance matrices, $\mat{Q}$ and $\mat{R}$. }
                \begin{tabular}{cc|cccc} \hline \hline
                    \multicolumn{2}{c|}{$\mat{Q}$ Parameters} &
                    \multicolumn{4}{c}{$\mat{R}$ Parameters}
                    \\ \hline
                    $\bv{}{}{\sigma}{2}{acc}$ & 
                    \multirow{2}{*}{$\bv{}{}{\sigma}{2}{qori}$} &
                    \multirow{2}{*}{$\bv{}{}{\sigma}{2}{ori}$} &
                    $\bv{}{}{\sigma}{2}{mp}$ &
                    $\bv{}{}{\sigma}{2}{ls}$ and $\bv{}{}{\sigma}{2}{rs}$ &
                    $\bv{}{}{\sigma}{2}{lim}$
                    \\
                    (m$^2$.s$^{-4}$) & 
                     &
                     &
                    (m$^2$) &
                    (m$^2$.s$^{-2}$ and m$^2$) &
                    (m$^2)$
                    \\ \hline
                    $10^2\mat{1}_{9}$ & 
                    $10^3\mat{1}_{12}$ &
                    $10^{-2}$ &
                    $0.1$ & 
                    $[0.01\mat{1}_{3} \: 10^{-4}]$ & 
                    $10\mat{1}_{18}$ \\
                    \hline \hline
                \end{tabular} 
                
                where $\mat{1}_{n}$ is an $1 \times n$ row vector with all elements equal to $1$.  
                \label{tab:var-param-kfnoise}
            \end{table}

        Lastly, the evaluation was done using the following metrics:
        (1) joint angles RMSE with bias removed and coefficient of correlation (CC) of the hip in the Y, X, and Z planes and of the knee in the Y plane;
        and (2) Total travelled distance (TTD) deviation (\ie{} TTD error with respect to the actual TTD) of the ankles.
        Refer to \cite[Sec. III]{sy2020ckf} for more details.
        
    \section{Results}
\begin{figure}[htbp]
            \centering
            \includegraphics[width=\linewidth]{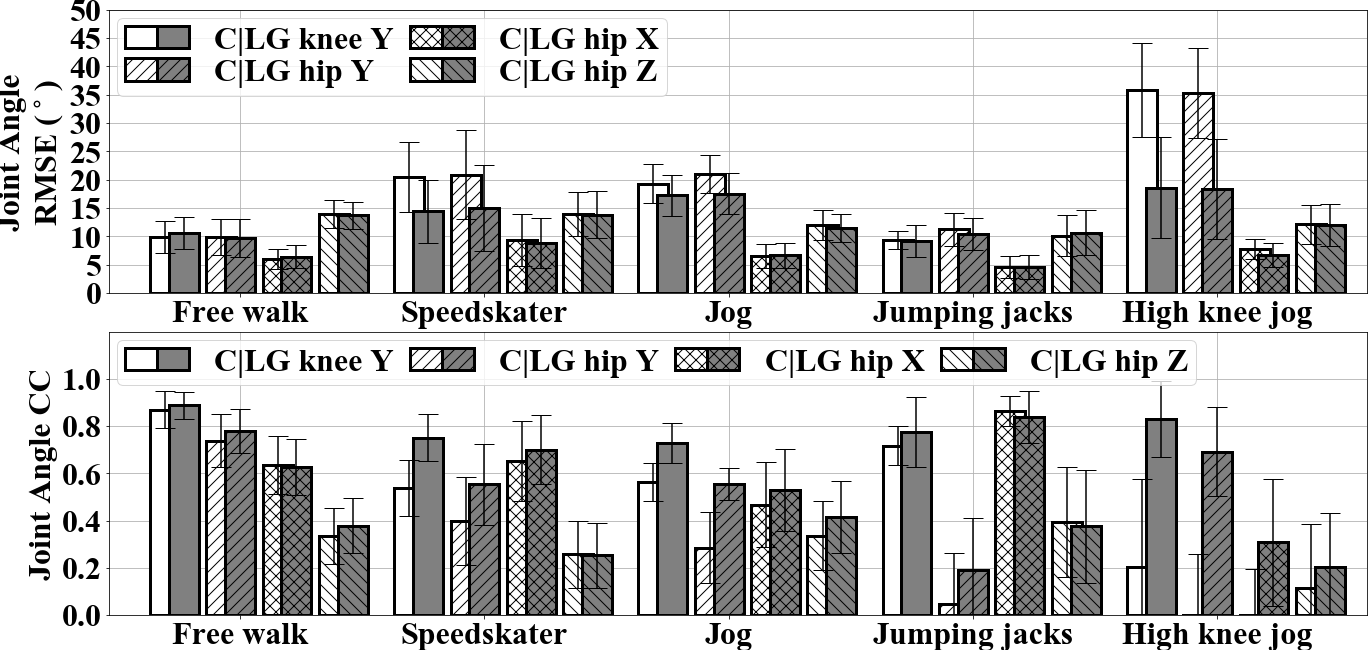}
            \caption{The CC of knee (Y) and hip (Y, X, Z) joint angles for \emph{LGKF-3IMU} (prefix \emph{LG}) and \emph{CKF-3IMU} (prefix \emph{C}) at each motion type.}
            \label{fig:results-kneehip-angles-rmsecc}
        \end{figure}

\begin{figure}[htbp]
            \centering
            \includegraphics[width=\linewidth]{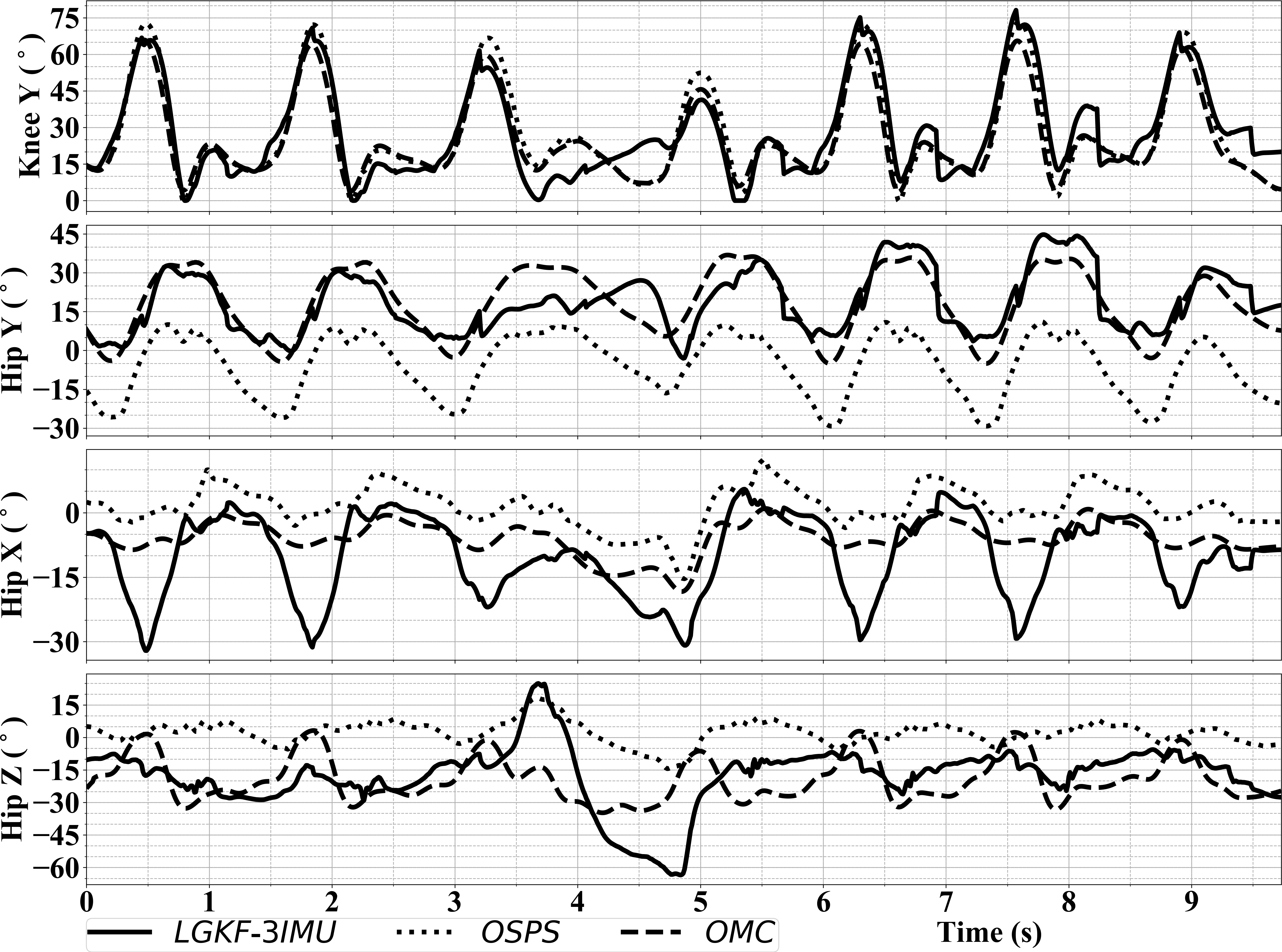}
            \caption{Knee (Y) and hip (Y, X, Z) joint angle output of \emph{LGKF-3IMU} in comparison with the benchmark system (Vicon) for a 
            \emph{Walk} trial. The subject walked straight from $t=0$ to $3$ s, turned $180^\circ$ around from $t=3$ to $5.5$ s, and walked straight to the original starting point from $5.5$ s until the end.}
            \label{fig:results-kneehip-angle-sample}
        \end{figure}

        Fig. \ref{fig:results-kneehip-angles-rmsecc} shows the knee and hip joint angle RMSE (bias removed) and CC compared against the OMC output.
            Y, X, and Z refers to the sagittal, frontal, and transverse planes, respectively.
        Fig. \ref{fig:results-kneehip-angle-sample} shows a sample \emph{Walk} trial.
        Table \ref{tab:results-ttddev} shows the TTD deviation at the ankles for free walk and jogging.
            Refer to \url{http://bit.ly/3bHlVG9} for video reconstructions of sample trials.
            
        \begin{table}[htbp]
            \centering
            \caption{Total travelled distance (TTD) deviation from optical motion capture (OMC) system at the ankles}
			\begin{tabular}{lrrrr}
				\hline
				\hline
				& \multicolumn{2}{c}{\emph{CKF-3IMU}} & \multicolumn{2}{c}{\emph{LGKF-3IMU}} \bigstrut[t]\\
				& \multicolumn{1}{c}{Left} & \multicolumn{1}{c}{Right} & \multicolumn{1}{c}{Left} & \multicolumn{1}{c}{Right} \bigstrut[b]\\
				\hline
				Free walk & 3.81\% & 3.61\% & 8.13\% & 8.13\% \bigstrut[t]\\
				Jog   & 24.05\% & 28.16\% & 18.58\% & 21.54\% \bigstrut[b]\\
				\hline
				\hline
			\end{tabular}%
            \label{tab:results-ttddev}%
        \end{table}%
	
\section{Discussion}
	    
    Fig. \ref{fig:results-kneehip-angles-rmsecc} shows that although there was minimal hip and knee joint angle RMSE and CC improvement for free walk between \emph{CKF-3IMU} and \emph{LGKF-3IMU}, 
        there was significant improvement for most dynamic movements, specifically, speedskater, jog, and high knee jog, 
        indicating that the Lie group representation has indeed made the pose estimator capable of tracking more ADLs and not just walking.
        This result also agrees with \cite{Joukov2017}.
    Similar to IMC based systems, \emph{LGKF-3IMU} also follows the trend of having sagittal (Y axis) joint angles similar to that captured by OMC systems ($0.89$ knee Y and $0.78$ hip Y CCs), 
        but with significant difference in frontal and transverse (X and Z axis) joint angles \cite{sy2020ckf, Cloete2008}.
        Similar qualitative observations can be seen in Fig. \ref{fig:results-kneehip-angle-sample}, specifically, there were larger angle change for hip X ($t=0$ to $3$ s and $t=6$ to $8$ s) and hip Z ($t=3$ to $5$ s).

    The knee and hip joint angle RMSEs and CCs of \emph{CKF-3IMU}, \emph{LGKF-3IMU}, \emph{OSPS} and related literature for free walking are shown in Table \ref{tab:knee-hip-angle-rmse-cc} \cite{sy2020ckf, Cloete2008}.
        Although the biased joint angle RMSE for \emph{LGKF-3IMU} is comparable with \emph{OSPS} and Cloete's ($<6^\circ$),
        the unbiased results show that utilizing fewer sensors does reduce accuracy somewhat \cite{Cloete2008}.
    Despite \emph{LGKF-3IMU} achieving good joint angle CCs in the sagittal plane, the unbiased joint angle RMSE ($>5^\circ$) makes its utility in clinical applications uncertain \cite{McGinley2009}.
        	Furthermore, \emph{LGKF-3IMU} shares the limitations of \emph{CKF-3IMU} during longer-term tracking of ADL, being unable to handle the activities of sitting, lying down, or climbing stairs due to the pelvis height and/or flat floor assumptions;
        	    and unable to track people with varus or valgus deformity, or those capable of hyperextending the knee due to the algorithm's hinge knee joint and ROM constraints.
            Developing solutions to further increase accuracy and overcome the said limitations (\eg{} measuring inter-sensor distance, incorporating dynamics in addition to kinematics, or leveraging long-term recordings and gait patterns) will be part of future work.
            
    \begin{table}[htbp]
        \centering
        \caption{Knee and hip angle RMSE (top) and CC (bottom) of \emph{CKF-3IMU}, \emph{OSPS}, and related literature}
        \begin{tabular}{lrrrrr}
        	\hline
        	\hline
        	\multicolumn{2}{p{5em}}{Joint Angle\newline{}RMSE ($^\circ$)} & \multicolumn{1}{c}{knee sagittal} & \multicolumn{1}{c}{hip sagittal} & \multicolumn{1}{c}{hip frontal} & \multicolumn{1}{c}{hip transverse} \bigstrut\\
        	\hline
        	\multicolumn{1}{p{3.5em}}{\multirow{3}[1]{*}{\parbox{3.5em}{\emph{CKF-}\newline{}\emph{3IMU}}}} & \multicolumn{1}{l}{biased} & $11.1 \pm 2.9$ & $11.8 \pm 3.2$ & $7.5 \pm 3.1$ & $17.5 \pm 4.7$ \bigstrut[t]\\
        	& \multicolumn{1}{l}{mean} & $-1.2 \pm 4.2$ & $-4.3 \pm 4.4$ & $-2.2 \pm 4.2$ & $-4.0 \pm 9.7$ \\
        	& \multicolumn{1}{l}{no bias} & $10.0 \pm 2.8$ & $9.9 \pm 3.1$ & $6.1 \pm 1.8$ & $13.9 \pm 2.4$ \\
        	\multicolumn{1}{p{3.5em}}{\multirow{3}[0]{*}{\parbox{3.5em}{\emph{LGKF-}\newline{}\emph{3IMU}}}} & \multicolumn{1}{l}{biased} & $13.9 \pm 4.5$ & $11.6 \pm 4.1$ & $8.9 \pm 4.2$ & $17.0 \pm 4.4$ \\
        	& \multicolumn{1}{l}{mean} & $8.1 \pm 4.8$ & $4.6 \pm 4.3$ & $-4.0 \pm 5.3$ & $-3.3 \pm 9.0$ \\
        	& \multicolumn{1}{l}{no bias} & $10.5 \pm 2.8$ & $9.7 \pm 3.3$ & $6.4 \pm 2.1$ & $13.7 \pm 2.4$ \\
        	\multirow{3}[0]{*}{\emph{OSPS}} & \multicolumn{1}{l}{biased} & $7.9 \pm 3.2$ & $12.4 \pm 6.0$ & $6.2 \pm 2.6$ & $19.8 \pm 6.6$ \\
        	& \multicolumn{1}{l}{mean} & $0.2 \pm 6.1$ & $-10.9 \pm 7.4$ & $0.2 \pm 2.5$ & $8.8 \pm 8.8$ \\
        	& \multicolumn{1}{l}{no bias} & $5.0 \pm 1.7$ & $3.6 \pm 1.7$ & $4.1 \pm 2.2$ & $11.9 \pm 4.3$ \\
        	\multicolumn{1}{p{3.5em}}{\multirow{2}[1]{*}{\parbox{3.6em}{Cloete \etal{}\cite{Cloete2008}}}} & \multicolumn{1}{l}{biased} & $11.5 \pm 6.4$ & $16.9 \pm 3.6$ & $9.6 \pm 5.1$ & $16.0 \pm 8.8$ \\
        	& \multicolumn{1}{l}{no bias} & $8.5 \pm 5.0$ & $5.8 \pm 3.8$ & $7.3 \pm 5.2$ & $7.9 \pm 4.9$ \bigstrut[b]\\
        	\hline
        	\hline
        	\multicolumn{2}{l}{Joint Angle CC} & \multicolumn{1}{c}{knee sagittal} & \multicolumn{1}{c}{hip sagittal} & \multicolumn{1}{c}{hip frontal} & \multicolumn{1}{c}{hip transverse} \bigstrut\\
        	\hline
        	\multicolumn{2}{l}{\emph{CKF-3IMU}} & $0.87 \pm 0.08$ & $0.74 \pm 0.11$ & $0.64 \pm 0.12$ & $0.33 \pm 0.12$ \bigstrut[t]\\
        	\multicolumn{2}{l}{\emph{LGKF-3IMU}} & $0.89 \pm 0.06$ & $0.78 \pm 0.09$ & $0.63 \pm 0.12$ & $0.38 \pm 0.12$ \\
        	\multicolumn{2}{l}{\emph{OSPS}} & $0.97 \pm 0.03$ & $0.95 \pm 0.06$ & $0.72 \pm 0.19$ & $0.26 \pm 0.20$ \\
        	\multicolumn{2}{l}{Cloete \etal \cite{Cloete2008}} & $0.89 \pm 0.15$ & $0.94 \pm 0.08$ & $0.55 \pm 0.40$ & $0.54 \pm 0.20$ \bigstrut[b]\\
        	\hline
        	\hline
        \end{tabular}%
      \label{tab:knee-hip-angle-rmse-cc}%
    \end{table}
        
	Comparing processing times, \emph{LG-CEKF} was slower than \emph{CKF} but can still be used in real time; 
	    specifically, \emph{LG-CEKF} and \emph{CKF} processed a 1,000-frame sequence in \texttildelow$2$ and \texttildelow$0.7$ seconds, respectively, on an Intel Core i5-6500 3.2 GHz CPU \cite{sy2020ckf}, 
		while the algorithm in \cite{Marcard2017} took 7.5 minutes on a quad-core Intel Core i7 3.5 GHz CPU. 
		All set-ups used single-core non-optimized Matlab code.
			
    Table \ref{tab:results-ttddev} shows that despite successful reconstruction of relative pose, \emph{LGKF-3IMU} had worse TTD for free walking than \emph{CKF-3IMU}.
        It can be observed from the sample video trial that the \emph{LGKF-3IMU} had less displacement during the turn around (\ie{} high rotational change).
    
	\emph{LGKF-3IMU} was able to achieve comparable and occasionally better results than \emph{CKF-3IMU} using 
	    fewer assumptions 
	        (\ie{} encourage pelvis $x$ and $y$ position to approach the average of the left and right ankle $x$ and $y$ positions during the measurement update, 
	        and the prevention of knee angle decrease during the constraint update \cite[Sec. II-E.2 and 3]{sy2020ckf});
        and only at one iteration (\emph{CKF-3IMU} used an iterative projection scheme called smoothly constrained KF),
	    indicating the robustness brought by the Lie group representation.
		Furthermore, \emph{LGKF-3IMU} does not assume perfect orientation during the constraint update, in contrast to \emph{CKF-3IMU}, which can be beneficial if new sensor information that informs segment orientation is added.
	
	\section{Conclusion}
    	This paper presented a Lie group CEKF-based algorithm (\emph{LGKF-3IMU}) to estimate lower limb kinematics using a reduced sensor count configuration, and without using any reference motion database.
    		The knee and hip joint angle RMSEs in the sagittal plane for free walking were $10.5 \pm 2.8^\circ$ and $9.7 \pm 3.3^\circ$, respectively, while the CCs were $0.89 \pm 0.06$ and $0.78 \pm 0.09$, respectively.
		We also showed that \emph{LGKF-3IMU} improves estimates for dynamic motion, and enables better convergence for our non-linear biomechanical constraints.
   		To further improve performance, additional information relating the pelvis and ankle kinematics is needed (\eg{} utilize sensors that give pelvis distance or position relative to the ankle). 
    	The source code for the \emph{LG-CEKF} algorithm, supplementary material, and links to sample videos will be made available at \url{https://git.io/Jv3oF}.
    	
    \addtolength{\textheight}{-12cm}
	\section*{Acknowledgement}
	    This research was supported by an Australian Government Research Training Program (RTP) Scholarship.
    
    \section*{References}
    \printbibliography[heading=none]

@Misc{		  sy2020lgcekfsupp,
  title		= {Supplementary material to Estimating Lower Limb Kinematics using a Lie Group Constrained EKF and a Reduced Wearable IMU Count},
  author	= {Sy, Luke and Lovell, Nigel H and
		  Redmond, Stephen J},
}

@Book{		  barfoot2017state,
  author	= {Barfoot, Timothy D},
  publisher	= {Cambridge University Press},
  title		= {{State Estimation for Robotics}},
  year		= {2017}
}

@Article{	  Bourmaud2014,
  author	= {Bourmaud, Guillaume and M{\'{e}}gret, R{\'{e}}mi and
		  Arnaudon, Marc and Giremus, Audrey},
  doi		= {10.1007/s10851-014-0517-0},
  issn		= {09249907},
  journal	= {J. Math. Imaging Vis.},
  number	= {1},
  pages		= {209--228},
  title		= {{Continuous-Discrete Extended Kalman Filter on Matrix Lie
		  Groups Using Concentrated Gaussian Distributions}},
  volume	= {51},
  year		= {2014}
}

@Book{		  Chirikjian2012Book2,
  doi		= {10.1007/978-0-8176-4944-9},
  title={Stochastic Models, Information Theory, and Lie Groups, Volume 2: Analytic Methods and Modern Applications},
  author={Chirikjian, Gregory S},
  volume={2},
  year={2011},
  publisher={Springer Science \& Business Media}
}

@InProceedings{	  Brossard2017,
  author	= {Brossard, Martin and Bonnabel, Silvere and Condomines,
		  Jean Philippe},
  booktitle	= {IEEE Int. Conf. Intell. Robot. Syst.},
  doi		= {10.1109/IROS.2017.8206066},
  isbn		= {9781538626825},
  issn		= {21530866},
  pages		= {2485--2491},
  title		= {{Unscented Kalman filtering on Lie groups}},
  volume	= {2017-Septe},
  year		= {2017}
}

@Article{	  DelRosario2018,
  author	= {{Del Rosario}, Michael B. and Ngo, Phillip and Khamis,
		  Heba and Lovell, Nigel H. and Redmond, Stephen J. and Ngo,
		  Phillip and Lovell, Nigel H. and Redmond, Stephen J.},
  doi		= {10.1109/JSEN.2018.2864989},
  issn		= {1530-437X},
  journal	= {IEEE Sens. J.},
  number	= {22},
  pages		= {9332--9342},
  publisher	= {IEEE},
  title		= {{Computationally efficient adaptive error-state Kalman
		  filter for attitude estimation}},
  volume	= {18},
  year		= {2018}
}

@Book{		  stillwell2008naive,
  author	= {Stillwell, John},
  publisher	= {Springer Science {\&} Business Media},
  title		= {{Naive lie theory}},
  year		= {2008}
}

@Article{	  Tautges2011,
    author	= {Tautges, Jochen and Zinke, A. and Kr{\"{u}}ger, B and
		  Baumann, J and Weber, Andreas and Helten, T and
		  M{\"{u}}ller, M and Seidel, H.P. and Eberhardt, B.},
  doi		= {10.1145/PREPRINT},
  eprint	= {1006.4903},
  isbn		= {0103660054},
  issn		= {07300301},
  journal	= {ACM Trans. Graph.},
  number	= {3},
  pages		= {18},
  title		= {{Motion reconstruction using sparse accelerometer data}},
  volume	= {30},
  year		= {2011}
}

@Article{	  Barfoot2014,
  author	= {Barfoot, Timothy D and Furgale, Paul T},
  doi		= {10.1109/TRO.2014.2298059},
  issn		= {15523098},
  journal	= {IEEE Trans. Robot.},
  number	= {3},
  pages		= {679--693},
  publisher	= {IEEE},
  title		= {{Associating uncertainty with three-dimensional poses for
		  use in estimation problems}},
  volume	= {30},
  year		= {2014}
}

@Article{	  DelRosario2016a,
  author	= {{Del Rosario}, Michael B. and Lovell, Nigel H. and
		  Redmond, Stephen J.},
  doi		= {10.1109/JSEN.2016.2574124},
  issn		= {1530437X},
  journal	= {IEEE Sens. J.},
  number	= {15},
  pages		= {6008--6017},
  title		= {{Quaternion-based complementary filter for attitude
		  determination of a smartphone}},
  volume	= {16},
  year		= {2016}
}

@Unpublished{	  sy2020ckf,
    author	= {Sy, Luke and Raitor, Michael and Rosario, Michael Del and
		  Khamis, Heba and Kark, Lauren and Lovell, Nigel H and
		  Redmond, Stephen J},
  eprint	= {1910.00910},
  primaryclass	= {cs.RO},
  title		= {{Estimating Lower Limb Kinematics using a Reduced Wearable
		  Sensor Count}},
  year		= {2020. To be published.}
}

@Article{	  Roetenberg2009,
  author	= {Roetenberg, Daniel and Luinge, Henk and Slycke, Per},
  doi		= {10.1.1.569.9604},
  isbn		= {9781595936349},
  journal	= {Xsens Motion Technol. BV, Tech. Rep},
  title		= {{Xsens MVN: Full 6DOF human motion tracking using
		  miniature inertial sensors}},
  volume	= {1},
  year		= {2009}
}

@Article{	  McGinley2009,
  author	= {McGinley, Jennifer L. and Baker, Richard and Wolfe, Rory
		  and Morris, Meg E.},
  doi		= {10.1016/j.gaitpost.2008.09.003},
  issn		= {09666362},
  journal	= {Gait Posture},
  number	= {3},
  pages		= {360--369},
  title		= {{The reliability of three-dimensional kinematic gait
		  measurements: A systematic review}},
  volume	= {29},
  year		= {2009}
}

@InCollection{	  selig2004lie,
  author	= {Selig, Jon M},
  booktitle	= {Comput. Noncommutative Algebr. Appl.},
  pages		= {101--125},
  publisher	= {Springer},
  title		= {{Lie groups and lie algebras in robotics}},
  year		= {2004}
}

@Article{	  Llorens2015,
  author	= {Llor{\'{e}}ns, Roberto and Gil-G{\'{o}}mez, Jos{\'{e}}
		  Antonio and Alca{\~{n}}iz, Mariano and Colomer, Carolina
		  and No{\'{e}}, Enrique},
  doi		= {10.1177/0269215514543333},
  isbn		= {1477-0873 (Electronic)$\backslash$r0269-2155 (Linking)},
  issn		= {14770873},
  journal	= {Clin. Rehabil.},
  number	= {3},
  pages		= {261--268},
  pmid		= {25056999},
  title		= {{Improvement in balance using a virtual reality-based
		  stepping exercise: A randomized controlled trial involving
		  individuals with chronic stroke}},
  volume	= {29},
  year		= {2015}
}

@InProceedings{	  Shull2010,
  author	= {Shull, Pete and Lurie, Kristen and Shin, Mihye and Besier,
		  Thor and Cutkosky, Mark},
  booktitle	= {2010 IEEE Haptics Symp.},
  doi		= {10.1109/HAPTIC.2010.5444625},
  isbn		= {9781424468218},
  issn		= {2324-7347},
  organization	= {IEEE},
  pages		= {409--416},
  title		= {{Haptic gait retraining for knee osteoarthritis
		  treatment}},
  year		= {2010}
}

@Article{	  Joukov2017,
  author	= {Joukov, Vladimir and Cesic, Josip and Westermann, Kevin
		  and Markovic, Ivan and Kulic, Dana and Petrovic, Ivan},
  doi		= {10.1109/IROS.2017.8206016},
  isbn		= {9781538626825},
  issn		= {21530866},
  journal	= {IEEE Int. Conf. Intell. Robot. Syst.},
  month		= {dec},
  pages		= {1965--1972},
  publisher	= {Institute of Electrical and Electronics Engineers Inc.},
  title		= {{Human motion estimation on Lie groups using IMU
		  measurements}},
  volume	= {2017-Septe},
  year		= {2017}
}

@Article{	  Joukov2019,
  author	= {Joukov, Vladimir and Cesic, Josip and Westermann, Kevin
		  and Markovic, Ivan and Petrovic, Ivan and Kulic, Dana},
  doi		= {10.1109/tcyb.2019.2933390},
  issn		= {2168-2267},
  journal	= {IEEE Trans. Cybern.},
  month		= {sep},
  pages		= {1--12},
  publisher	= {Institute of Electrical and Electronics Engineers (IEEE)},
  title		= {{Estimation and Observability Analysis of Human Motion on
		  Lie Groups}},
  year		= {2019}
}

@Article{	  Lin2012,
  author	= {Lin, Jonathan F.S. and Kuli{\'{c}}, Dana},
  doi		= {10.1088/0967-3334/33/12/2099},
  issn		= {09673334},
  journal	= {Physiol. Meas.},
  number	= {12},
  pages		= {2099--2115},
  title		= {{Human pose recovery using wireless inertial measurement
		  units}},
  volume	= {33},
  year		= {2012}
}

@Article{	  Meng2012,
  author	= {Meng, X L and Zhang, Z Q and Sun, S Y and Wu, J K and
		  Wong, W C},
  doi		= {10.1088/0957-0233/23/5/055101},
  issn		= {0957-0233},
  journal	= {Meas. Sci. Technol.},
  month		= {may},
  number	= {5},
  pages		= {055101},
  publisher	= {IOP Publishing},
  title		= {{Biomechanical model-based displacement estimation in
		  micro-sensor motion capture}},
    volume	= {23},
  year		= {2012}
}

@Article{	  Wang2006,
  author	= {Wang, Yunfeng and Chirikjian, Gregory S.},
  doi		= {10.1109/TRO.2006.878978},
  issn		= {15523098},
  journal	= {IEEE Trans. Robot.},
  number	= {4},
  pages		= {591--602},
  title		= {{Error propagation on the Euclidean group with
		  applications to manipulator kinematics}},
  volume	= {22},
  year		= {2006}
}

@Article{	  Bourmaud2013,
  author	= {Bourmaud, Guillaume and Giremus, Audrey and Berthoumieu,
		  Yannick and Bourmaud, Guillaume},
  pages		= {1--5},
  title		= {{Discrete extended Kalman filter on lie groups}},
  year		= {2013}
}

@Article{	  Cesic2016,
  author	= {{\'{C}}esi{\'{c}}, Josip and Joukov, Vladimir and
		  Petrovi{\'{c}}, Ivan and Kuli{\'{c}}, Dana},
  doi		= {10.1109/HUMANOIDS.2016.7803369},
  isbn		= {9781509047185},
  issn		= {21640580},
  journal	= {IEEE-RAS Int. Conf. Humanoid Robot.},
  pages		= {826--833},
  title		= {{Full body human motion estimation on lie groups using 3D
		  marker position measurements}},
  year		= {2016}
}

@InProceedings{	  Cloete2008,
  author	= {Cloete, Teunis and Scheffer, Cornie},
  booktitle	= {2008 30th Annu. Int. Conf. IEEE Eng. Med. Biol. Soc.},
  doi		= {10.1109/IEMBS.2008.4650232},
  isbn		= {978-1-4244-1814-5},
  month		= {aug},
  pages		= {4579--4582},
  publisher	= {IEEE},
  title		= {{Benchmarking of a full-body inertial motion capture
		  system for clinical gait analysis}},
    year		= {2008}
}

@InProceedings{	  Marcard2017,
    author	= {von Marcard, Timo and Rosenhahn, Bodo and Black, Michael J
		  and Pons-Moll, Gerard},
  booktitle	= {Comput. Graph. Forum},
  doi		= {10.1111/cgf.13131},
  eprint	= {1703.08014},
  issn		= {14678659},
  number	= {2},
  organization	= {Wiley Online Library},
  pages		= {349--360},
  title		= {{Sparse inertial poser: Automatic 3D human pose estimation
		  from sparse IMUs}},
  volume	= {36},
  year		= {2017}
}

@InProceedings{	  Huang2018a,
    author	= {Huang, Yinghao and Kaufmann, Manuel and Aksan, Emre and
		  Black, Michael J. and Hilliges, Otmar and Pons-Moll,
		  Gerard},
  booktitle	= {SIGGRAPH Asia 2018 Tech. Pap. SIGGRAPH Asia 2018},
  doi		= {10.1145/3272127.3275108},
  eprint	= {1810.04703},
  isbn		= {9781450360081},
  issn		= {15577368},
  month		= {dec},
  publisher	= {Association for Computing Machinery, Inc},
  title		= {{Deep inertial poser: Learning to reconstruct human pose
		  from sparse inertial measurements in real time}},
  year		= {2018}
}
    
\end{document}


\maketitle

	Section \ref{sec:mathbg} will give a mathematical background of Lie group and Lie algebra.
		Useful properties of Lie groups relevant to 3D pose estimation (i.e., $SO(3)$, $SE(3)$, and $\R^n$) will also be listed.
	Section \ref{sec:systemmodel}-\ref{sec:const-update} will then describe our algorithm indepth.
		
\section{Mathematical Background} \label{sec:mathbg}
    Sec. \ref{sec:mathbg-lg} will give a mathematical background of Lie group and Lie algebra.
		Sec. \ref{sec:mathbg-so3}, \ref{sec:mathbg-se3}, \ref{sec:mathbg-R} will define useful properties for the special orthogonal group, $SO(3)$, special euclidean group, $SE(3)$, and vectors, $\R^n$, respectively.
		
\subsection{Lie group and Lie algebra} \label{sec:mathbg-lg}
	The matrix Lie group $\LG{G}$ is a group of $n \times n$ matrices that is also a smooth manifold (\eg{} $\LG{SE(3)}$).
    	Group composition and inversion (\ie{} matrix multiplication and inversion) are smooth operations.
	Lie algebra $\LA{g}$ represents a tangent space of a group at the identity element \cite{selig2004lie}.
	    The elegance of Lie theory lies in it being able to represent curved objects using a vector space (\eg{} Lie group $\LG{G}$ represented by $\LA{g}$) \cite{stillwell2008naive}.
	        	
	The matrix exponential $\exp{}_{\LG{G}}: \LA{g} \tiny{\to} \LG{G}$ and matrix logarithm $\log{}_{\LG{G}}: \LG{G} \tiny{\to} \LA{g}$ establish a local diffeomorphism between the Lie group $\LG{G}$ and its Lie algebra $\LA{g}$.
        Eq. \eqref{eq:expG} shows the definition of $\exp{}_{\LG{G}}$ where $\Lhat{}{\mat{\phi}} \in \LA{g}$.
	    The Lie algebra $\LA{g}$ is a $n \times n$ matrix that can be represented compactly with an $n$ dimensional vector space. A linear isomorphism between $\LA{g}$ and $\R^n$ is given by
            $\Lvee{G}{\:\:}: \LA{g} \tiny{\to} \R^n$ and
            $\Lhat{G}{\:\:}: \R^n \tiny{\to} \LA{g}$.
        An illustration of the said mappings are given in Fig. \ref{fig:lie-group-algebra-overview}.
    Furthermore, the adjoint operators of a Lie group, denoted as $\LAdSmall{G}{X}$; 
        the adjoint operators of a Lie algebra, denoted as $\LadSmall{G}{X}$;
        and the right jacobian, denoted as $\LJac{G}{\vec{v}}$, will be used in later sections.
        For a more detailed introduction to Lie groups refer to \cite{barfoot2017state, Chirikjian2012Book2}. 
	        For an accessible introduction to Lie theory, refer to \cite{stillwell2008naive}.
	
    \begin{figure}
        \centering
        \resizebox{0.4\linewidth}{!}{
		\begin{tikzpicture}
			\begin{scope}
			\clip (-2.5,0) rectangle (2.5,2.5);
			\draw (0,0) circle (2.5) node[above left] (A) {Lie group $\LG{G}$};
			\end{scope}
			\draw[<->] (1.77-1.5,1.77+1.5) -- (1.77+1.5,1.77-1.5)
			node[pos=0.5, label={above right:{Lie algebra $\LA{g}$}}] (B) {};
			\draw[<->] (3.5,2.5) -- (5.5,2.5)
			node[pos=0.7, above] (C) {$\R^n$};
			\draw[->] (A) .. controls +(up:1cm) and +(left:1cm) .. (B)
			node[pos=0.5, above]{$\log{}_{\LG{G}}$};
			\draw[->] (B) .. controls +(down:1cm) and +(right:2cm) .. (A)
			node[pos=0.5, below right]{$\exp{}_{\LG{G}}$};
			\draw[->] (B) .. controls +(up:2cm) and +(up:1cm) .. (C)
			node[pos=0.5, below]{$\Lvee{G}{\:\:}$};
			\draw[->] (C) .. controls +(down:2cm) and +(right:1cm) .. (B)
			node[pos=0.5, below]{$\Lhat{G}{\:\:}$};
		\end{tikzpicture}}
        \caption{Mapping between Lie group $\LG{G}$, Lie algebra $\LA{g}$, and a $n$-dimensional vector space.}
        \label{fig:lie-group-algebra-overview}
    \end{figure}
	
	\begin{align}
		\exp{}_{\LG{G}} \left(\Lhat{}{\mat{\phi}}\right) = \sum_{n=0}^{\infty} \frac{1}{n!} \left( \Lhat{}{\mat{\phi}} \right)^n \label{eq:expG} \\
		\LJac{G}{\vec{v}} = \sum_{i=0}^{\infty} \frac{(-1)^i}{(i+1)!} \Lad{G}{\vec{v}}^i \text{ , } \vec{v} \in \R^p
	\end{align}

\subsection{Special Orthogonal Group $SO(3)$} \label{sec:mathbg-so3}
	The Special Orthogonal Group $SO(3)$ represents orientation.
	Note that $\mat{C}$ is the typical rotation matrix $\R^{3 \times 3}$ whose column vectors represent the $x$, $y$, and $z$ basis vectors.
	\begin{align}
		\LG{SO(3)} := \left\{ \mat{C} \in \R^{3 \times 3} | \mat{C}\mat{C}^T = 1, \det \mat{C}=1 \right\}
	\end{align}
	
	The basic operations for $\LG{SO(3)}$ are listed below.
		See \cite[Ch. 7]{barfoot2017state} for details.
	
	\begin{gather}
		\Lhat{SO(3)}{\mat{\phi}} = \Lhat{SO(3)}{\begin{matrix} \phi_1 \\ \phi_2 \\ \phi_3 \end{matrix}} 
			= \begin{bmatrix}
				0 & -\phi_3 & \phi_2 \\
				\phi_3 & 0 & -\phi_1 \\
				-\phi_2 & \phi_1 & 0
			\end{bmatrix} \\
		\Lvee{SO(3)}{\mat{C}} = \Lvee{SO(3)}{\begin{matrix}
			0 & -\phi_3 & \phi_2 \\
			\phi_3 & 0 & -\phi_1 \\
			-\phi_2 & \phi_1 & 0
			\end{matrix}} 
		= \begin{bmatrix} \phi_1 \\ \phi_2 \\ \phi_3 \end{bmatrix} \\
		\Lvectran{SO(3)}{\mat{\phi}} = \cos \left( |\mat{\phi}| \right) \mat{I}_{3 \times 3} + (1 - \cos\left( |\mat{\phi}| \right) ) \frac{\mat{\phi} \mat{\phi}^T}{ |\mat{\phi}|^2 } + \sin \left( |\mat{\phi}| \right) \Lhat{SO(3)}{ \frac{\mat{\phi}}{|\mat{\phi}|} } \\
		\log{}_{\LG{SO(3)}} \left( \mat{X} \right) = \frac{\theta}{2 \sin(\theta)} \left( \mat{X} - \mat{X}^T \right) \text{ s.t. } 1 + 2 \cos(\theta) = \Tr(\mat{X})
		\begin{cases}
			\theta \ne o & -\pi < 0 < \pi \\
			\theta = 0 & \log \left( \mat{X} \right) = 0
		\end{cases} \\
		\LAd{SO(3)}{\mat{X}} = \mat{X} \\
		\Lad{SO(3)}{\mat{x}} = \Lhat{SO(3)}{\mat{x}} \\
		\mat{X}^{-1} = \mat{X}^T
	\end{gather}
		
\subsection{Special Euclidean Group $SE(3)$} \label{sec:mathbg-se3}
	The Special Euclidean Group $SE(3)$ represents translation and orientation.
	\begin{align}
		\LG{SE(3)} := \left\{ \mat{T}=\begin{bmatrix}
		\mat{C} & \mat{r} \\ \mat{0}^T & 1
		\end{bmatrix} \in \R^{4 \times 4} | \left\{ \mat{C}, \mat{r} \right\} \in \LG{SO(3)} \times \R^3 \right\}
	\end{align}

	The basic operations for $\LG{SE(3)}$ are listed below.
		See \cite[Ch. 7]{barfoot2017state} for details.
	\begin{gather}
		\Lhat{SE(3)}{\mat{\xi}} = \Lhat{SE(3)}{\begin{matrix} \vec{\rho} \\ \vec{\phi} \end{matrix}} 
		= \begin{bmatrix}
		\Lhat{SO(3)}{\vec{\phi}} & \vec{\rho} \\
		0 & 0
		\end{bmatrix} \\
		\Lvee{SE(3)}{\mat{X}} = \Lvee{SO(3)}{\begin{matrix}
				\mat{C} & \vec{r} \\
				0 & 0
			\end{matrix}} 
		= \begin{bmatrix} \Lvee{SO(3)}{\mat{C}} \\ \vec{r} \end{bmatrix} \\
		\Lvectran{SE(3)}{\mat{\xi}} = \Lvectran{SE(3)}{ \begin{matrix} \vec{\rho} \\ \vec{\phi} \end{matrix} }
			= \begin{bmatrix}
				\exp_{\LG{SO(3)}} \left( \Lhat{SO(3)}{\vec{\phi}} \right) & \mat{J} \vec{\rho} \\
				0 & 1
			\end{bmatrix} \\
		\mat{J} = \frac{\sin \left( |\mat{\phi}| \right)}{|\mat{\phi}|} \mat{I}_{3 \times 3} + \left(1 - \frac{\sin \left( |\mat{\phi}| \right)}{|\mat{\phi}|} \right) \frac{\mat{\phi} \mat{\phi}^T}{ |\mat{\phi}|^2 } + \frac{1 - \cos( |\vec{\phi}| )}{|\vec{\phi}|} \Lhat{SO(3)}{ \frac{\mat{\phi}}{|\mat{\phi}|} } \\
		\LAd{SE(3)}{\mat{X}}
			= \LAd{SE(3)}{\begin{bmatrix}
					\mat{C} & \vec{r} \\
					0 & 0
				\end{bmatrix}}
			= \begin{bmatrix}
				\mat{C} & \vec{r} \mat{C} \\
				\mat{0} & \mat{C}
			\end{bmatrix} \\
		\Lad{SE(3)}{\vec{\xi}} = \Lad{SE(3)}{\begin{bmatrix} \vec{\rho} \\ \vec{\phi} \end{bmatrix}} 
			= \begin{bmatrix}
				\Lhat{SO(3)}{\vec{\phi}} & \Lhat{SO(3)}{\vec{\rho}} \\
				\mat{0} & \Lhat{SO(3)}{\vec{\phi}}
			\end{bmatrix} \\
		\mat{X}^{-1} = \begin{bmatrix}
				\mat{C} & \vec{r} \\
				0 & 0
			\end{bmatrix}^{-1} 
			= \begin{bmatrix}
				\mat{C}^T & -\mat{C}^T \vec{r} \\
				\mat{0} & 0
			\end{bmatrix}^{-1} 
	\end{gather}
	
	Also note of another useful property as shown below for $\vec{a}, \vec{b} \in \LA{se(3)}$.
	    This property will be used later on when differentiating the equations for the biomechanical constraints.
		See \cite[][Eq. (72)]{barfoot2017state} for details.
	
	\begin{gather}
		\Lhat{\LG{SE(3)}}{\bv{}{}{a}{}{}} \bv{}{}{b}{}{} = \bv{}{}{a}{}{} \Lodot{\LG{SE(3)}}{\bv{}{}{b}{}{}} \label{eq:se3-swap} \\
		\begin{bmatrix}
			\epsilon \\ \eta
		\end{bmatrix}^\odot = \begin{bmatrix}
			\eta \mat{I}_{3 \times 3} & -\Lhat{SO(3)}{\epsilon} \\
			\mat{0}_{1 \times 3} & \mat{0}_{1 \times 3}
		\end{bmatrix}
	\end{gather}

\subsection{Real numbers $\R^n$} \label{sec:mathbg-R}
	Real numbers $\R^n$ can represent translation, velocity, acceleration, and any other states that can be represented as vectors.
	The basic Lie group and algebra operations for $\R^n$ are listed below.
	
	\begin{gather}
	\Lhat{\R^n}{\vec{v}} = \begin{bmatrix}
	\mat{0}_{n \times n} & \vec{v} \\
	\mat{0}_{1 \times n} & 0 
	\end{bmatrix} \\
	\Lvee{\R^n}{\mat{X}} = \Lvee{\R^n}{\begin{matrix}
		\mat{0}_{n \times n} & \vec{v} \\
		\mat{0}_{1 \times n} & 0 
		\end{matrix}} 
	= \vec{v} \\
	\exp{}_{\LG{\R^n}} \left( \Lhat{\R^n}{\vec{v}} \right) = \begin{bmatrix}
		\mat{I}_{n \times n} & \vec{v} \\
		\mat{0}_{1 \times n} & 1
	\end{bmatrix} \\
	\log{}_{\LG{\R^n}} \left( \mat{X} \right) = \log{}_{\LG{\R^n}} \left( \begin{bmatrix}
			\mat{I}_{n \times n} & \vec{v} \\
			\mat{0}_{1 \times n} & 1
			\end{bmatrix} \right)
		= \begin{bmatrix}
			\mat{0}_{n \times n} & \vec{v} \\
			\mat{0}_{1 \times n} & 0 
		\end{bmatrix} \\
	\LAd{\R^n}{\mat{X}} = \mat{I}_{n \times n} \\
	\Lad{\R^n}{\mat{x}} = \mat{0}_{n \times n} \\
	\mat{X_1}\mat{X_2} = \exp{}_{\LG{\R^n}} \left( \Lhat{\R^n}{\vec{v_1}} \right) \exp{}_{\LG{\R^n}} \left( \Lhat{\R^n}{\vec{v_1}} \right) = \exp{}_{\LG{\R^n}} \left( \Lhat{\R^n}{\vec{v_1} + \vec{v_2}} \right) \\
	\mat{X}^{-1} = \begin{bmatrix}
			\mat{I}_{n \times n} & \vec{v} \\
			\mat{0}_{1 \times n} & 1
		\end{bmatrix}^{-1} 
		= \begin{bmatrix}
			\mat{I}_{n \times n} & -\vec{v} \\
			\mat{0}_{1 \times n} & 1
		\end{bmatrix}
	\end{gather}
	
\section{System, measurement, and constraint models} \label{sec:systemmodel}
  	The system and measurement models are presented below
	    \begin{gather}
	        \vec{X}_{k} = f(\vec{X}_{k\smallneg1}, \vec{n}_{k\smallneg1}) = \vec{X}_{k\smallneg1} \exp_{G} ( \Lhat{G}{\Omega (\vec{X}_{k\smallneg1} ) \tiny{+} \vec{n}_{k\smallneg1}} ) \label{eq:pred-update} \\
	        \vec{Z}_{k} = h (\vec{X}_{k}) \exp_{G}\left(\Lhat{G}{\vec{m_{k}}}\right)  ,\:\:
	        \vec{D}_{k} = c (\vec{X}_{k}) \label{eq:meas-cstr-update}
        \end{gather}
    	
	where 
	\begin{itemize}
		\item $k$ is the time step;
		\item $\vec{X}_{k} \in \LG{G}$ is the system state, an element of state Lie group $\LG{G}$;
		\item $\Omega \left(\vec{X}_{k}\right) : \LG{G} \tiny{\to} \R^p$ is a non-linear function;
		\item $\vec{n}_{k}$ is a zero-mean process noise vector with covariance matrix $\mat{Q}_{k}$ (\ie{} $\vec{n}_k \sim \N_{\R^p}(\vec{0}_{p \times 1}, \mat{Q}_{k})$);
		\item $\vec{Z}_{k} \in \LG{G_1}$ is the system measurement, an element of measurement Lie group $\LG{G_1}$;
		\item $ h\left(\vec{X}_{k}\right): \LG{G} \tiny{\to} \LG{G_1}$ is the measurement function;
		\item $\vec{m}_{k}$ is a zero-mean measurement noise vector with covariance matrix $\mat{R}_{k}$ (\ie{} $\vec{m}_k \sim \N_{\R^q}(\vec{0}_{q \times 1}, \mat{R}_{k})$);
		\item $\mat{D}_{k} \in \LG{G_2}$ is the constraint state,
    	 an element of constraint Lie group $\LG{G_2}$;
		\item $ c\left(\vec{X}_{k}\right): \LG{G} \tiny{\to} \LG{G_2}$ is the equality constraint function the state $\vec{X}_{k}$ must satisfy.
	\end{itemize}
	
    Similar to \cite{Bourmaud2013, Cesic2016}, the state distribution of $\vec{X}_{k}$ is assumed to be a concentrated Gaussian distribution on Lie groups (\ie{} $\vec{X}_{k} = \vec{\mu}_k \exp_{\LG{G}} \Lhat{G}{\vec{\epsilon}}$ where $\vec{\mu}_k$ is the mean of $\vec{X}_{k}$ and Lie algebra error $\vec{\epsilon} \sim \N_{\R^p}(\vec{0}_{p \times 1}, \mat{P})$) \cite{Wang2006}.
    The Lie group state variables $\vec{X}_{k}$ model the position, orientation, and velocity of the three instrumented body segments (\ie{} pelvis and shanks) as shown in Eq. \eqref{eq:Lgroup-state} (\ie{} $\LG{G} = \LG{SE(3)}^3 \times \R^{9}$).
    	$\bv{A}{B}{T}{}{} \in \LG{SE(3)}$ denotes the pose of body segment $B$ relative to frame $A$.
    	    If frame $\cs{A}$ is not specified, assume reference to the world frame, $\cs{W}$.
	Other basic Lie group operators for $\LG{G}$ are explicitly listed below where $\Lhat{}{\bv{W}{B}{\xi}{}{}}$ is the Lie algebra of $\bv{W}{B}{T}{}{}$ for some body segment $B$.

	\begin{gather}
		\mat{X}_{k} = \begin{bmatrix}
				\bv{W}{p}{T}{}{} & & & \\
				& \bv{W}{ls}{T}{}{} & & \\
				& & \bv{W}{rs}{T}{}{} & \\
				& & & \Lvectran{\R^n}{\begin{matrix}
					\bv{W}{p}{v}{}{} \\ \bv{W}{ls}{v}{}{} \\ \bv{W}{rs}{v}{}{}
					\end{matrix}} \\
			\end{bmatrix} 
			= \begin{bmatrix}
				\bv{W}{p}{T}{}{} & & & & & & \\
				 & \bv{W}{ls}{T}{}{} & & & & & \\
				 & & \bv{W}{rs}{T}{}{} & & & & \\
				 & & & \mat{I}_{3 \times 3} & & & \bv{W}{p}{v}{}{} \\
				 & & & & \mat{I}_{3 \times 3} & & \bv{W}{ls}{v}{}{} \\
				 & & & & & \mat{I}_{3 \times 3} & \bv{W}{rs}{v}{}{} \\
				 & & & & & & 1\\
			\end{bmatrix} \label{eq:Lgroup-state} \\
		\Lhat{G}{\mat{x}_k} = \Lhat{G}{\begin{matrix}
				\bv{W}{p}{\xi}{}{} \\ \bv{W}{ls}{\xi}{}{} \\ \bv{W}{rs}{\xi}{}{} \\
				\bv{W}{p}{v}{}{} \\ \bv{W}{ls}{v}{}{} \\ \bv{W}{rs}{v}{}{}
			\end{matrix}}
			= \begin{bmatrix}
				\Lhat{SE(3)}{\bv{W}{p}{\xi}{}{}} & & & \\
				& \Lhat{SE(3)}{\bv{W}{ls}{\xi}{}{}} & & \\
				& & \Lhat{SE(3)}{\bv{W}{rs}{\xi}{}{}} & \\
				& & & \Lhat{\R^n}{\begin{matrix}
					\bv{W}{p}{v}{}{} \\
					\bv{W}{ls}{v}{}{} \\
					\bv{W}{rs}{v}{}{}
					\end{matrix}} \\
			\end{bmatrix} \label{eq:Lhat-def}\\
		\Lvee{G}{\mat{Y}_k} = \Lvee{G}{\begin{matrix}
			\bv{W}{p}{Y}{}{} & & & \\
			& \bv{W}{ls}{Y}{}{} & & \\
			& & \bv{W}{rs}{Y}{}{} & \\
			& & & \bv{W}{vel}{Y}{}{} \end{matrix}}
			= \begin{bmatrix}
			\Lvee{SE(3)}{\bv{W}{p}{Y}{}{}} & & & \\
			& \Lvee{SE(3)}{\bv{W}{ls}{Y}{}{}} & & \\
			& & \Lvee{SE(3)}{\bv{W}{rs}{Y}{}{}} & \\
			& & & \Lvee{\R^n}{\bv{W}{vel}{Y}{}{}} \\
			\end{bmatrix} \\
		\begin{split}
		\Lvectran{G}{\mat{x}_k} &= \Lvectran{G}{\begin{matrix}
				\bv{W}{p}{\xi}{}{} \\ \bv{W}{ls}{\xi}{}{} \\ \bv{W}{rs}{\xi}{}{} \\
				\bv{W}{p}{v}{}{} \\ \bv{W}{ls}{v}{}{} \\ \bv{W}{rs}{v}{}{}
			\end{matrix}} \\
			&= \begin{bmatrix}
				\Lvectran{SE(3)}{\bv{W}{p}{\xi}{}{}} & & & \\
				& \Lvectran{SE(3)}{\bv{W}{ls}{\xi}{}{}} & & \\
				& & \Lvectran{SE(3)}{\bv{W}{rs}{\xi}{}{}} & \\
				& & & \Lvectran{\R^n}{\begin{matrix}
					\bv{W}{p}{v}{}{} \\ \bv{W}{ls}{v}{}{} \\ \bv{W}{rs}{v}{}{}
					\end{matrix} } \\
			\end{bmatrix}
		\end{split} \\
		\Ltranvec{G}{\mat{X}_k} = \Ltranvec{G}{\begin{bmatrix}
			\bv{W}{p}{T}{}{} & & & \\
			& \bv{W}{ls}{T}{}{} & & \\
			& & \bv{W}{rs}{T}{}{} & \\
			& & & \Lvectran{\R^n}{\begin{matrix}
				\bv{W}{p}{v}{}{} \\ \bv{W}{ls}{v}{}{} \\ \bv{W}{rs}{v}{}{}
				\end{matrix}} \end{bmatrix}}
			= \begin{bmatrix}
				\Ltranvec{SE(3)}{\bv{W}{p}{T}{}{}} \\
				\Ltranvec{SE(3)}{\bv{W}{ls}{T}{}{}} \\
				\Ltranvec{SE(3)}{\bv{W}{rs}{T}{}{}} \\
				\bv{W}{p}{v}{}{} \\ \bv{W}{ls}{v}{}{} \\ \bv{W}{rs}{v}{}{}
			\end{bmatrix} \\
			\LAd{G}{\mat{X}_k} = \begin{bmatrix}
					\LAd{SE(3)}{\bv{W}{p}{T}{}{}} & & & \\
					& \LAd{SE(3)}{\bv{W}{ls}{T}{}{}} & & \\
					& & \LAd{SE(3)}{\bv{W}{rs}{T}{}{}} & \\
					& & & \mat{I}_{9 \times 9} 
				\end{bmatrix} \\
			\Lad{}{\mat{x}_k} = \Lad{}{\begin{bmatrix}
				\bv{W}{p}{\xi}{}{} \\ \bv{W}{ls}{\xi}{}{} \\ \bv{W}{rs}{\xi}{}{} \\
				\bv{W}{p}{v}{}{} \\ \bv{W}{ls}{v}{}{} \\ \bv{W}{rs}{v}{}{}
				\end{bmatrix}} 
				= \begin{bmatrix}
					\Lad{SE(3)}{\bv{W}{p}{\xi}{}{}} & & & \\
					& \Lad{SE(3)}{\bv{W}{ls}{\xi}{}{}} & & \\
					& & \Lad{SE(3)}{\bv{W}{rs}{\xi}{}{}} & \\
					& & & \mat{0}_{9 \times 9} 
				\end{bmatrix} \\
			\LJac{G}{\mat{v}} = \begin{bmatrix}
				\LJac{SE(3)}{\bv{W}{p}{T}{}{}} & & & \\
				& \LJac{SE(3)}{\bv{W}{ls}{T}{}{}} & & \\
				& & \LJac{SE(3)}{\bv{W}{rs}{T}{}{}} & \\
				& & & \mat{I}_{9 \times 9} 
			\end{bmatrix}
	\end{gather}
			
\section{Prediction update} \label{sec:pred-update}
	Below are the Lie group EKF \textit{a-priori} state and state error covariance matrix propagation equations as defined in \cite{Bourmaud2013}.
	Note that measured acceleration and orientation are denoted as $\bvmeas{W}{B}{a}{}{k}$ and $\bvmeas{W}{B}{R}{}{k}$, respectively, for segment $B$.
	
    \begin{gather}
    	\kfsp{\mu}{k+1} = \kfsc{\mu}{k} \exp_{G} ( 
							\LhatSmall{G}{\kfsc{\Omega}{k}} 
   						) \label{eq:lgkf-predmu} \\
	    \kfcp{P}{k+1} = \mathcal{F}_{k} \kfcm{P}{k} \mathcal{F}_{k}^T + \LJacSmall{G}{\hat{\Omega}_k} \mat{Q}_{k} \LJacSmall{G}{\hat{\Omega}_k}^T \label{eq:lgkf-predP} \\
	    \mathcal{F}_{k} = \LAdSmall{G}{\exp_{G} ( -\LhatSmall{G}{\hat{\Omega}_k} )} + \LJacSmall{G}{\hat{\Omega}_k} \mathscr{C}_k \\
	    \mathscr{C}_{k} = \tfrac{\partial}{\partial \vec{\epsilon}} 
	    \Omega\left( \vec{\mu}_{k}^\epsilon \right) |_{\vec{\epsilon} = 0}
    \end{gather}
	where 
	\begin{gather}
		\mu_{k}^\epsilon = \mu_{k} \Lvectran{G}{\vec{\epsilon}} \\
		\vec{\epsilon} = \begin{bmatrix}
				\bv{W}{p}{\epsilon}{T}{\rho} &
				\bv{W}{p}{\epsilon}{T}{\phi} &
				\bv{W}{ls}{\epsilon}{T}{\rho} &
				\bv{W}{ls}{\epsilon}{T}{\phi} &
				\bv{W}{rs}{\epsilon}{T}{\rho} &
				\bv{W}{rs}{\epsilon}{T}{\phi} &
				\bv{W}{p}{\epsilon}{T}{v} &
				\bv{W}{ls}{\epsilon}{T}{v} &
				\bv{W}{rs}{\epsilon}{T}{v}
			\end{bmatrix}^T \\
		\mat{Q}_{k} = \diag \left(
			\tfrac{\dt^2}{2} \bv{}{mp}{\sigma}{}{acc},
			\bv{}{mp}{\sigma}{}{qori},
			\tfrac{\dt^2}{2} \bv{}{ls}{\sigma}{}{acc},
			\bv{}{ls}{\sigma}{}{qori},
			\tfrac{\dt^2}{2} \bv{}{rs}{\sigma}{}{acc},
			\bv{}{rs}{\sigma}{}{qori},
			\dt \bv{}{mp}{\sigma}{}{acc},
			\dt \bv{}{ls}{\sigma}{}{acc},
			\dt \bv{}{rs}{\sigma}{}{acc} \right) \\
		\kfsc{\Omega}{k} = \Omega \left( \kfsc{\mu}{k} \right) \\
		\Omega \left( \mat{X}{k} \right) = \begin{bmatrix}
			\bvmeas{W}{p}{R}{T}{k} \left(
			\dt \bv{W}{mp}{v}{}{k} +
			\tfrac{\dt^2}{2} \bvmeas{W}{p}{a}{}{k} 
			\right) \\
			\bv{}{}{0}{}{3 \times 1} \\
			\bvmeas{W}{ls}{R}{T}{k} \left(
			\dt \bv{W}{la}{v}{}{k} +
			\tfrac{\dt^2}{2} \bvmeas{W}{ls}{a}{}{k} 
			\right) \\
			\bv{}{}{0}{}{3 \times 1} \\
			\bvmeas{W}{rs}{R}{T}{k} \left(
			\dt \bv{W}{ra}{v}{}{k} +
			\tfrac{\dt^2}{2} \bvmeas{W}{rs}{a}{}{k} 
			\right) \\
			\bv{}{}{0}{}{3 \times 1} \\
			\dt \bvmeas{W}{mp}{a}{}{k} \\
			\dt \bvmeas{W}{la}{a}{}{k} \\
			\dt \bvmeas{W}{ra}{a}{}{k} 
		\end{bmatrix}
	\end{gather}
To calculate for $\mathscr{C}_k$,
	\begin{gather}
		\Omega\left( \mu_{k}^\epsilon \right) = \begin{bmatrix}
				\bvmeas{W}{p}{R}{T}{k} \left(
				\dt (\bv{W}{mp}{v}{}{k} + \bv{W}{p}{\epsilon}{}{v} ) +
				\tfrac{\dt^2}{2} \bvmeas{W}{p}{a}{}{k} 
				\right) \\
				\bv{}{}{0}{}{3 \times 1} \\
				\bvmeas{W}{ls}{R}{T}{k} \left(
				\dt (\bv{W}{la}{v}{}{k} + \bv{W}{la}{\epsilon}{}{v} ) +
				\tfrac{\dt^2}{2} \bvmeas{W}{ls}{a}{}{k} 
				\right) \\
				\bv{}{}{0}{}{3 \times 1} \\
				\bvmeas{W}{rs}{R}{T}{k} \left(
				\dt (\bv{W}{ra}{v}{}{k} + \bv{W}{ra}{\epsilon}{}{v} ) +
				\tfrac{\dt^2}{2} \bvmeas{W}{rs}{a}{}{k} 
				\right) \\
				\bv{}{}{0}{}{3 \times 1} \\
				\dt \bvmeas{W}{mp}{a}{}{k} \\
				\dt \bvmeas{W}{la}{a}{}{k} \\
				\dt \bvmeas{W}{ra}{a}{}{k}
			\end{bmatrix} \\[1.5em]
			\mathscr{C}_k = \left[ \begin{array}{c:ccc}
					\bovermat{SE(3)}{\multirow{6}{*}{$\mat{0}_{18\times18}$}} &
					\bovermat{vel}{\dt \bvmeas{W}{p}{R}{T}{k} & \quad\mat{0}_{3 \times 3}\quad & \quad\mat{0}_{3 \times 3} \quad} \\
					& \mat{0}_{3 \times 3} & \mat{0}_{3 \times 3} & \mat{0}_{3 \times 3} \\
					& \mat{0}_{3 \times 3} & \dt \bvmeas{W}{ls}{R}{T}{k} & \mat{0}_{3 \times 3} \\
					& \mat{0}_{3 \times 3} & \mat{0}_{3 \times 3} & \mat{0}_{3 \times 3} \\
					& \mat{0}_{3 \times 3} & \mat{0}_{3 \times 3} & \dt \bvmeas{W}{rs}{R}{T}{k} \\
					& \mat{0}_{3 \times 3} & \mat{0}_{3 \times 3} & \mat{0}_{3 \times 3} \\ \hdashline
					\multicolumn{4}{c}{\raisebox{0pt}{$\mat{0}_{9 \times 27}$}} \\
				\end{array} \right]
	\end{gather}
	With these equations, we should have all the ingredients defined for the prediction update.
		
\section{Measurement update} \label{sec:meas-update}
	The \textit{a posteriori} state $\kfsm{\mu}{k}$ is calculated following the Lie EKF equations below.
		See \cite{Bourmaud2013} for more details.
	\begin{gather}
		\mat{K}_{k} = \kfcp{P}{k} \mathcal{H}_{k}^T \left( \mathcal{H}_{k} \kfcp{P}{k} \mathcal{H}_{k}^T + \mat{R}_{k} \right)^{-1} \\
		\vec{\nu}_{k} = \mat{K}_{k} \left( \Ltranvec{G_1}{ h(\kfsp{\mu}{k})^{-1} \mat{Z}_k } \right) \\
		\mathcal{H}_{k} = \frac{\partial}{\partial \vec{\epsilon}}
			\overbrace{
			\Ltranvec{G_1}{
					h\left( \kfsp{\mu}{k} \right)^{-1} 
					h\left( \vec{\mu}_k^\epsilon \right) }
			}^{\mat{\delta h}} \bigg|_{\vec{\epsilon}=0}  \\
		\kfsm{\mu}{k} = \kfsp{\mu}{k} \Lvectran{G}{\vec{\nu}_k}
	\end{gather}
	where $\mat{R}_{k} = \diag(\vec{\sigma}_{k})$.
	
	As floor contact (FC) varies with time, $\mathcal{H}_{k}$ varies with time as shown in Eq. \eqref{eq:H-k-cases}.	
		Measurement variance  $\bv{}{}{\sigma}{}{k}$ is constructed similarly to Eq. \eqref{eq:H-k-cases}.
	$\mat{Z}_k$ is shown in Eq. \eqref{eq:Z-k-cases}.
		$h \left( \mat{X}_k \right)$ is constructed similarly to Eq. \eqref{eq:Z-k-cases}.
	\begin{gather}
		\mathcal{H}_{k} = \begin{cases}
		[ \mathcal{H}_{ori}^T \quad \mathcal{H}_{mp}^T ]^T & \text{ no FC} \\
		[ \mathcal{H}_{ori}^T \quad \mathcal{H}_{mp}^T \quad \mathcal{H}_{ls}^T ]^T & \text{ left FC} \\
		[ \mathcal{H}_{ori}^T \quad \mathcal{H}_{mp}^T \quad \mathcal{H}_{rs}^T ]^T & \text{ right FC} \\
		[ \mathcal{H}_{ori}^T \quad \mathcal{H}_{mp}^T \quad \mathcal{H}_{ls}^T \quad \mathcal{H}_{rs}^T ]^T & \text{ both FC} \\
		\end{cases} \label{eq:H-k-cases} \\
		\mat{Z}_{k} = \begin{cases}
		\diag( \mat{Z}_{ori}^T , \mat{Z}_{mp}^T ) & \text{ no FC} \\
		\diag( \mat{Z}_{ori}^T , \mat{Z}_{mp}^T , \mat{Z}_{ls}^T ) & \text{ left FC} \\
		\diag( \mat{Z}_{ori}^T , \mat{Z}_{mp}^T , \mat{Z}_{rs}^T ) & \text{ right FC} \\
		\diag( \mat{Z}_{ori}^T , \mat{Z}_{mp}^T , \mat{Z}_{ls}^T , \mat{Z}_{rs}^T ) & \text{ both FC} \\
		\end{cases} \label{eq:Z-k-cases}
	\end{gather}
	
	Below are important notes to make the derivation of $\mathcal{H}$ easier:
  \begin{enumerate}
      \item If $h_a \left( \mat{X}_k \right) \in \R^p$ instead of $\in SE(3)$ for some measurement type $a$ with $p$ dimensions, then $\mat{X}_1^{-1} \mat{X}_2 = \mat{X}_2 - \mat{X}_1$.
      \item It follows from above that $\mat{\delta h}_a =
			\Ltranvec{G_a}{ 
				h_a \left( \kfsp{\mu}{k} \right)^{-1} 
				h_a \left( \vec{\mu}_k^\epsilon \right) } = h\left( \vec{\mu}_k^\epsilon \right) - h\left( \kfsp{\mu}{k} \right)$ where $\LG{G_a}$ is some Lie group representing the measurement space of $a$.
  \end{enumerate}
	
\subsection{Orientation update}
 Firstly, the orientation update is implemented by $\mathcal{H}_{ori}$ and $\mat{Z}_{ori,k}$ as shown in Eqs. \eqref{eq:H-ori-k} and \eqref{eq:Z-ori-k} with measurement noise variance $\bv{}{}{\sigma}{2}{ori}$ ($9 \times 1$ vector).

 \begin{gather}
  \mat{Z}_{ori} = \diag(\bvmeas{W}{p}{R}{}{k}, \bvmeas{W}{ls}{R}{}{k}, \bvmeas{W}{rs}{R}{}{k}) \label{eq:Z-ori-k} \\
  h_{ori} \left( \mat{X}_{k} \right) = \diag(\bv{W}{p}{R}{}{k}, \bv{W}{ls}{R}{}{k}, \bv{W}{rs}{R}{}{k})
 		\end{gather}
 		\begin{align}
    	\mat{\delta h}_{ori} &= \Ltranvec{G_{ori}}{ 
    		h_{ori} \left( \kfsp{\mu}{k} \right)^{-1} 
    		h_{ori} \left( \vec{\mu}_k^\epsilon \right) } \\
   		&= \Ltranvec{G_{ori}}{ 
   			\diag(\bv{W}{p}{\bar{R}}{T}{k}\bv{W}{p}{\bar{R}}{}{k} \exp_{}(\bv{W}{p}{\epsilon}{}{\phi}),
 				  \bv{W}{ls}{\bar{R}}{T}{k}\bv{W}{ls}{\bar{R}}{}{k} \exp_{}(\bv{W}{ls}{\epsilon}{}{\phi}),
 				  \bv{W}{rs}{\bar{R}}{T}{k}\bv{W}{rs}{\bar{R}}{}{k} \exp_{}(\bv{W}{rs}{\epsilon}{}{\phi}) ) } \\
   		&= \Ltranvec{G_{ori}}{ 
			\diag \left(
			\LvectranSmall{SO(3)}{\bv{W}{p}{\epsilon}{}{\phi}}, \LvectranSmall{SO(3)}{\bv{W}{ls}{\epsilon}{}{\phi}},
		  	\LvectranSmall{SO(3)}{\bv{W}{rs}{\epsilon}{}{\phi}} \right) } \\
   		&= \begin{bmatrix}
    		\bv{W}{p}{\epsilon}{}{\phi} \\
    		\bv{W}{ls}{\epsilon}{}{\phi} \\
    		\bv{W}{rs}{\epsilon}{}{\phi}
   		\end{bmatrix} \\
    	\mathcal{H}_{ori} &= \left[ \begin{array}{cccccc:c}
    \mat{0}_{3 \times 3} & \mat{I}_{3 \times 3} & & & & & \\
    & & \mat{0}_{3 \times 3} & \mat{I}_{3 \times 3} & & & \mat{0}_{9 \times 9} \\
    & & & & \mat{0}_{3 \times 3} & \mat{I}_{3 \times 3} & \\
   \end{array} \right] \label{eq:H-ori-k}
 \end{align}
 
\subsection{Pelvis height assumption}
	Secondly, we assume the pelvis $z$ position to be close to the initial pelvis $z$ position as time $k=0$ (\ie{} standing height $z_{p}$).
		These constraints are implemented by $\mat{Z}_{mp}$ and $\mathcal{H}_{mp}$ as shown in Eqs. \eqref{eq:Z-mp-k} and \eqref{eq:H-mp-k} with measurement noise variance $\bv{}{}{\sigma}{2}{mp}$ ($1 \times 1$ vector).
		
	\begin{gather}
		\mat{Z}_{mp} = z_{p} \label{eq:Z-mp-k} \\
		h_{mp} \left( \mat{X}_{k} \right) = \vec{i}_z^T \bv{W}{p}{T}{}{} \vec{i}_0 \\
		\vec{i}_z = \begin{bmatrix} 0 & 0 & 1 & 0  \end{bmatrix}^T \\
		\vec{i}_0 = \begin{bmatrix} 0 & 0 & 0 & 1 \end{bmatrix}^T
	\end{gather}
	
	\begin{align}
		\mat{\delta h}_{mp} &= \Ltranvec{G_{mp}}{ 
			h_{mp} \left( \kfsp{\mu}{k} \right)^{-1} 
			h_{mp} \left( \vec{\mu}_k^\epsilon \right) } \quad \text{ Note that the output is $\in \R^3$}  \\
			&= h_{mp} \left( \vec{\mu}_k^\epsilon \right) - h_{mp} \left( \kfsp{\mu}{k} \right) \\
			&= \vec{i}_z^T \bv{W}{p}{\bar{T}}{}{} \Lvectran{SE(3)}{\LAtwobare{W}{p}{\epsilon}{}} \vec{i}_0 - \vec{i}_z^T \bv{W}{p}{\bar{T}}{}{} \vec{i}_0 \\
			& \text{ Linearize } \exp(\vec{\phi}) \approx \mat{I} + \Lhat{}{\vec{\phi}} \nonumber \\
			&= \vec{i}_z^T \bv{W}{p}{\bar{T}}{}{} \Lhat{}{\LAtwobare{W}{p}{\epsilon}{}} \vec{i}_0 \\
			&= \vec{i}_z^T \bv{W}{p}{\bar{T}}{}{} \Lodot{}{\vec{i}_0} \LAtwo{W}{p}{\epsilon}{} \quad \text{ Use Eq. \eqref{eq:se3-swap} to swap error state to the right} \\
		\mathcal{H}_{mp} &= \left[ \begin{array}{ccc:c}
			\vec{i}_z^T \bv{W}{p}{\bar{T}}{}{} \Lodot{}{\vec{i}_0} & \mat{0}_{1 \times 6} & \mat{0}_{1 \times 6} & \mat{0}_{1 \times 9} \\
			\end{array} \right] \label{eq:H-mp-k}
	\end{align}
	
\subsection{Zero velocity and flat floor update}
	Thirdly, if a left step is detected, the left ankle velocity is encouraged to approach zero, and the left ankle $z$ position to be close to the floor level, $z_{f}$.
		These assumptions are implemented using $\mat{Z}_{ls}$ and $\mathcal{H}_{ls}$ and  as shown in Eq. \eqref{eq:Z-lstep-k} and \eqref{eq:H-lstep-k} with measurement noise variance $\bv{}{}{\sigma}{2}{ls}$ ($4 \times 1$ vector).
		Note that $\mat{Z}_{rs}$ and $\mathcal{H}_{rs}$ can be constructed in a similar fashion to Eqs. \eqref{eq:Z-lstep-k} and \eqref{eq:H-lstep-k}.

 		\begin{gather}
		\mat{Z}_{ls} = \begin{bmatrix} \mat{0}_{1 \times 3} & z_{f} \end{bmatrix}^T \label{eq:Z-lstep-k} \\
		h_{ls} \left( \mat{X}_{k} \right) = \begin{bmatrix} 
				\bv{W}{ls}{v}{}{} \\
				\vec{i}_z^T \bv{W}{ls}{T}{}{} \vec{i}_0
			\end{bmatrix}
		\end{gather}
		\begin{align}
			\mat{\delta h}_{ls} &= \Ltranvec{G_{ls}}{ 
					h_{ls} \left( \kfsp{\mu}{k} \right)^{-1} 
					h_{ls} \left( \vec{\mu}_k^\epsilon \right) 
				} \quad \text{ Note that the output is $\in \R^4$}  \\
				& = h_{ls} \left( \vec{\mu}_k^\epsilon \right) - h_{ls} \left( \kfsp{\mu}{k} \right) \\
				& = \begin{bmatrix} 
						\bv{W}{ls}{\bar{v}}{}{} + \bv{W}{ls}{\epsilon}{}{v} - \bv{W}{ls}{\bar{v}}{}{} \\
						\vec{i}_z^T \bv{W}{ls}{\bar{T}}{}{} \Lvectran{}{\LAtwobare{W}{ls}{\epsilon}{}} \vec{i}_0 - \vec{i}_z^T \bv{W}{ls}{\bar{T}}{}{} \vec{i}_0
					\end{bmatrix} \\
				& \text{ Linearize } \exp(\vec{\phi}) \approx \mat{I} + \Lhat{}{\vec{\phi}} \nonumber \\
				& = \begin{bmatrix} 
						\bv{W}{ls}{\epsilon}{}{v} \\
						\vec{i}_z^T \bv{W}{ls}{\bar{T}}{}{} \Lhat{}{\LAtwobare{W}{ls}{\epsilon}{}} \vec{i}_0
					\end{bmatrix} \\
				& = \begin{bmatrix} 
						\bv{W}{ls}{\epsilon}{}{v} \\
						\vec{i}_z^T \bv{W}{ls}{\bar{T}}{}{} \Lodot{}{\vec{i}_0} \LAtwo{W}{ls}{\epsilon}{}
					\end{bmatrix} \quad \text{ Use Eq. \eqref{eq:se3-swap} to swap error state to the right} \\[1.5em]
			\mathcal{H}_{ls} &= \left[ \begin{array}{c:c:c:c:c}
					\multirow{2}{*}{$\dots$} & \bovermat{ls $\LG{SE(3)$} col.}{\qquad\qquad} & \multirow{2}{*}{$\dots$} & 
						\bovermat{vel. col.}{\mat{I}_{3 \times 3}} & \multirow{2}{*}{$\dots$} \\
					& \vec{i}_z^T \bv{W}{ls}{\bar{T}}{}{} \Lodot{}{\vec{i}_0} & & & \\
				\end{array} \right] \label{eq:H-lstep-k}
		\end{align}

\subsection{Covariance limiter}
	Lastly, the covariance limiter prevents the covariance from growing indefinitely and from becoming badly conditioned, as will happen naturally with the KF tracking the global position of the pelvis and ankles without any global position reference.
		At this step, a pseudo-measurement equal to the current state $\kfsm{\mu}{k}$ is used (implemented by $\mathcal{H}_{lim}$ as shown in Eq. \eqref{eq:H-lim})  with some measurement noise of variance $\bv{}{}{\sigma}{2}{lim}$ ($9 \times 1$ vector).
		The covariance $\kfcm{P}{k}$ is then calculated through Eqs. \eqref{eq:Hkprime}-\eqref{eq:Ptildekprime}.
	\begin{align}
		\mat{\delta h}_{lim} &= \Ltranvec{G_{lim}}{ 
				h_{lim} \left( \kfsp{\mu}{k} \right)^{-1} 
				h_{lim} \left( \vec{\mu}_k^\epsilon \right) 
			} \\
			&= \begin{bmatrix}
				\bv{W}{p}{\epsilon}{T}{\rho} &
				\bv{W}{p}{\epsilon}{T}{\phi} &
				\bv{W}{ls}{\epsilon}{T}{\rho} &
				\bv{W}{ls}{\epsilon}{T}{\phi} &
				\bv{W}{rs}{\epsilon}{T}{\rho} &
				\bv{W}{rs}{\epsilon}{T}{\phi}
			\end{bmatrix}^T \\
		\mathcal{H}_{lim} &= \begin{bmatrix} \mat{I}_{18 \times 18} & \mat{0}_{18 \times 18} \end{bmatrix} \label{eq:H-lim}
	\end{align}
	\begin{gather}
		\mathcal{H}_{k}' = \begin{bmatrix} \mathcal{H}_{k}^T & \mathcal{H}_{lim}^T \end{bmatrix}^T, \quad
		\mat{R}_{k}' = \diag([ \bv{}{}{\sigma}{}{k} \: \bv{}{}{\sigma}{}{lim} ]) \label{eq:Hkprime} \\
		\mat{K}_{k}' = \kfcp{P}{k} \mathcal{H}_{k}'^T \left( \mathcal{H}_{k}' \kfcp{P}{k} \mathcal{H}_{k}'^T + \mat{R}' \right)^{-1} \label{eq:Kkprime} \\
		\kfcm{P}{k} = \LJac{G}{\vec{\mu}_k} \left(\mat{I} - \mat{K}_{k}' \mathcal{H}_{k}' \right) \kfcp{P}{k} \LJac{G}{\vec{\nu}_k}^T \label{eq:Ptildekprime}
	\end{gather}
		
\section{Constraint update} \label{sec:const-update}
	The constrained state $\kfsc{\mu}{k}$ can be calculated using the equations below.
		Note $\mathcal{C}_{k} = \begin{bmatrix} \mathcal{C}_{L, k}^T & \mathcal{C}_{R, k}^T \end{bmatrix}^T$. 
		$\mathcal{C}_{L, k}$ is shown in Eq. \eqref{eq:D-left}. 
		Note that $\mathcal{C}_{R, k}$ can be derived similarly.
		Each of the constraints will be defined in the following subsections.
		\begin{align}
			\mathcal{C}_{L, k} &= \begin{cases}
			[ \mathcal{C}_{ltl,k}^T \quad \mathcal{C}_{lkh,k}^T ]^T 
			& \alpha_{lk,min} \leq \alpha_{lk} \leq \alpha_{lk,max}, \\
			[ \mathcal{C}_{ltl,k}^T \quad \mathcal{C}_{lkh,k}^T \quad \mathcal{C}_{lkr,k}^T ]^T & \text{otherwise.}
			\end{cases} \label{eq:D-left}
		\end{align}
		\begin{gather}
			\mat{K}_{k} = \kfcm{P}{k} \mathcal{C}_{k}^T \left( \mathcal{C}_{k} \kfcm{P}{k} \mathcal{C}_{k}^T \right)^{-1} \\
			\vec{\nu}_{k} = \mat{K}_{k} \left( \Lvee{G_2}{\log_{G_2} \left( c(\kfsm{\mu}{k})^{-1} \mat{D}_k \right) } \right) \\
			\mathcal{C}_{k} = \frac{\partial}{\partial \vec{\epsilon}} \Lvee{G_2}{
				\log_{G_2} \big( 
				c\left( \kfsm{\mu}{k} \right)^{-1} 
				c\left( \kfsm{\mu}{k} \exp_{G} \left( \Lhat{G}{\vec{\epsilon}} \right) \right) 
				\big) } |_{\vec{\epsilon}=0} \\
			\kfsc{\mu}{k} = \kfsm{\mu}{k} \exp_{G} \left( \Lhat{G}{\vec{\nu}_k} \right)
		\end{gather}

\subsection{Thigh length}
	Firstly, the constraint for the length of the estimated thigh vector is shown in Eq. \eqref{eq:c-lthigh} where $\bv{W}{lt}{\tau}{}{z}(\kfsc{\mu}{k})$ is the thigh vector (Eq. \eqref{eq:thigh-vect}, a function of the estimated state variables, $\kfsc{\mu}{k}$).
		The right thigh length constraint is derived similarly.
	\begin{gather}
		\mat{E} = \begin{bmatrix}
				\mat{I}_{3 \times 3} & \mat{0}_{3 \times 1}
			\end{bmatrix} \quad
		\bv{p}{lh}{p}{}{} = \begin{bmatrix}
				0 & \tfrac{d^{\cs{p}}}{2} & 0 & 1
			\end{bmatrix}^T \quad
		\bv{ls}{lk}{p}{}{} = \begin{bmatrix}
				0 & 0 & d^{\cs{ls}} & 1
			\end{bmatrix}^T \\
		\bv{W}{lt}{\tau}{}{z}(\kfsc{\mu}{k}) = 
			\overbrace{\mat{E} \bv{W}{p}{T}{}{} \bv{p}{lh}{p}{}{} }^{\text{hip joint pos.}} - 
			\overbrace{\mat{E} \bv{W}{ls}{T}{}{} \bv{ls}{lk}{p}{}{} }^{\text{knee joint pos.}} \label{eq:thigh-vect} \\
		\begin{split}
		c_{ltl}(\kfsc{\mu}{k}) &= \bv{W}{lt}{\tau}{}{z}(\kfsc{\mu}{k})^T 			
			\bv{W}{lt}{\tau}{}{z}(\kfsc{\mu}{k}) - (d^{\cs{lt}})^2 \label{eq:c-lthigh} \\
				&= 0 = \mat{D}_{ltl}
		\end{split}
	\end{gather}
	\begin{align}
		\mat{\delta c}_{ltl} &= \Ltranvec{G_{ltl}}{ 
				c_{ltl} \left( \kfsm{\mu}{k} \right)^{-1} 
				c_{ltl} \left( \vec{\mu}_k^\epsilon \right) 
			} \quad \text{ Note that the output is $\in \R^1$}  \\
			& = c_{ltl} \left( \vec{\mu}_k^\epsilon \right) - c_{ltl} \left( \kfsp{\mu}{k} \right) \\
			& \text{For simplicity let us define } \bv{W}{lt}{\tau}{}{z} \left( \vec{\mu}_k^\epsilon \right) and \bv{W}{lt}{\tau}{}{z} \left( \kfsm{\mu}{k} \right) \nonumber \\
		\bv{W}{lt}{\tau}{}{z} \left( \vec{\mu}_k^\epsilon \right) & = \mat{E} \left( 
				\bv{W}{p}{\bar{T}}{}{} \Lvectran{}{\LAtwo{W}{p}{\epsilon}{}} \bv{p}{lh}{p}{}{} - \bv{W}{ls}{\bar{T}}{}{} \Lvectran{}{\LAtwo{W}{ls}{\epsilon}{}} \bv{ls}{lk}{p}{}{}
			\right) \\
			& \text{ Linearize } \exp(\vec{\phi}) \approx \mat{I} + \Lhat{}{\vec{\phi}} \nonumber \\
			& = \mat{E} \left( 
				\bv{W}{p}{\bar{T}}{}{} \bv{p}{lh}{p}{}{} 
				- \bv{W}{ls}{\bar{T}}{}{} \bv{ls}{lk}{p}{}{}
				+ \bv{W}{p}{\bar{T}}{}{} \Lhat{}{\LAtwo{W}{p}{\epsilon}{}} \bv{p}{lh}{p}{}{}
				- \bv{W}{ls}{\bar{T}}{}{} \Lhat{}{\LAtwo{W}{ls}{\epsilon}{}} \bv{ls}{lk}{p}{}{}
			\right) \\
			& = \mat{E} \big( 
			\overbrace{
					\bv{W}{p}{\bar{T}}{}{} \bv{p}{lh}{p}{}{} 
					- \bv{W}{ls}{\bar{T}}{}{} \bv{ls}{lk}{p}{}{}
			}^{\mat{A}}
			+ \overbrace{
				\bv{W}{p}{\bar{T}}{}{} \Lodot{}{\bv{p}{lh}{p}{}{}} \LAtwo{W}{p}{\epsilon}{}
				- \bv{W}{ls}{\bar{T}}{}{} \Lodot{}{\bv{ls}{lk}{p}{}{}} \LAtwo{W}{ls}{\epsilon}{}
			}^{\mat{B}}
			\big) \\
		\bv{W}{lt}{\tau}{}{z} \left( \kfsm{\mu}{k} \right) &= \mat{E} \mat{A} \\
			& \text{Resuming the calculation for } \mat{\delta c}_{ltl} \nonumber \\
		\mat{\delta c}_{ltl} &= \left( \mat{A} + \mat{B} \right)^T \mat{E}^T \mat{E} \left( \mat{A} + \mat{B} \right) 
			- (d^{\cs{lt}})^2 
			- \left( \mat{A} \right)^T \mat{E}^T \mat{E} \left( \mat{A} \right) 
			+ (d^{\cs{lt}})^2 \\
			& \text{Note that } \mat{A}^T \mat{E}^T \mat{E} \mat{B} = \mat{B}^T \mat{E}^T \mat{E} \mat{A} \text{ since it is scalar} \nonumber \\
			&= \mat{A}^T \mat{E}^T \mat{E} \mat{A} + 2 \mat{A}^T \mat{E}^T \mat{E} \mat{B} + \mat{B}^T \mat{E}^T \mat{E} \mat{B} - \mat{A}^T \mat{E}^T \mat{E} \mat{A} \\
			& \text{Assume second order error } \mat{B}^T \mat{E}^T \mat{E} \mat{B} \approx 0 \nonumber \\
			&= 2 \mat{A}^T \mat{E}^T \mat{E} \mat{B} \\
			&= 2 \mat{A}^T \mat{E}^T \mat{E} \left(
				\bv{W}{p}{\bar{T}}{}{} \Lodot{}{\bv{p}{lh}{p}{}{}} \LAtwo{W}{p}{\epsilon}{}
				- \bv{W}{ls}{\bar{T}}{}{} \Lodot{}{\bv{ls}{lk}{p}{}{}} \LAtwo{W}{ls}{\epsilon}{}
			\right) \\
		\mathcal{C}_{ltl,k} &= \left[ \begin{array}{ccc:c}
			2 \mat{A}^T \mat{E}^T \mat{E} \bv{W}{p}{\bar{T}}{}{} \Lodot{}{\bv{p}{lh}{p}{}{}} & 
			-2 \mat{A}^T \mat{E}^T \mat{E} \bv{W}{ls}{\bar{T}}{}{} \Lodot{}{\bv{ls}{lk}{p}{}{}} & 
			\mat{0}_{1 \times 6} & \mat{0}_{1 \times 9} \\
		\end{array} \right] \label{eq:C-ltl-k}
	\end{align}
	
\subsection{Hinge knee joint}
	Secondly, the constraint for the hinge knee joint enforces the long ($z$) axis of the thigh to be perpendicular to the mediolateral axis ($y$) of the shank, as shown in Eq. \eqref{eq:c-lkhinge}.
		This formulation is similar to \cite[][Sec. 2.3 Eqs. (4)]{Meng2012}.

	\begin{align}
		c_{lkh}(\vec{\mu}_k) &= \bv{W}{ls}{r}{T}{y} \bv{W}{lt}{\tau}{}{z} \label{eq:c-lkhinge} \\
			&= \left( \mat{E} \bv{W}{ls}{T}{}{} \vec{i}_{y} \right)^T \bv{W}{lt}{\tau}{}{z} \\
			&= 0 = \mat{D}_{lkh}
	\end{align}
	\begin{align}
		\mat{\delta c}_{lkh} &= \Ltranvec{G_{lkh}}{ 
			c_{lkh} \left( \kfsm{\mu}{k} \right)^{-1} 
			c_{lkh} \left( \vec{\mu}_k^\epsilon \right) 
		} \quad \text{ Note that the output is $\in \R^1$}  \\
			& = c_{lkh} \left( \vec{\mu}_k^\epsilon \right) - c_{lkh} \left( \kfsp{\mu}{k} \right) \\
			&= \left( \mat{E} \bv{W}{ls}{\bar{T}}{}{} \Lvectran{}{\LAtwo{W}{ls}{\epsilon}{}} \vec{i}_{y} \right)^T \mat{E} \left( \mat{A} + \mat{B} \right) 
				- \left( \mat{E} \bv{W}{ls}{\bar{T}}{}{} \vec{i}_{y} \right)^T \mat{E} \mat{A} \\
			& \text{ Linearize } \exp(\vec{\phi}) \approx \mat{I} + \Lhat{}{\vec{\phi}} \nonumber \\
			&= \left( \mat{E} \left( \bv{W}{ls}{\bar{T}}{}{}  + \bv{W}{ls}{\bar{T}}{}{} \Lhat{}{\LAtwo{W}{ls}{\epsilon}{}} \right) \vec{i}_{y} \right)^T \mat{E} \left( \mat{A} + \mat{B} \right) 
			- \left( \mat{E} \bv{W}{ls}{\bar{T}}{}{} \vec{i}_{y} \right)^T \mat{E} \mat{A} \\
			&= \left( \mat{E} \bv{W}{ls}{\bar{T}}{}{} \vec{i}_{y} \right)^T \mat{E} \mat{A} 
			+ \left( \mat{E} \bv{W}{ls}{\bar{T}}{}{}  \vec{i}_{y} \right)^T \mat{E} \mat{B} 
			+ \left( \mat{E} \bv{W}{ls}{\bar{T}}{}{} \Lhat{}{\LAtwo{W}{ls}{\epsilon}{}} \vec{i}_{y} \right)^T \mat{E} \mat{A} \nonumber \\
			&\quad + \left( \mat{E} \bv{W}{ls}{\bar{T}}{}{} \Lhat{}{\LAtwo{W}{ls}{\epsilon}{}} \vec{i}_{y} \right)^T \mat{E} \mat{B} 
			- \left( \mat{E} \bv{W}{ls}{\bar{T}}{}{} \vec{i}_{y} \right)^T \mat{E} \mat{A} \\
			& \text{Assume second order error } \approx 0 \nonumber \text{ and scalar so transposable} \\
			&= \left( \mat{E} \bv{W}{ls}{\bar{T}}{}{}  \vec{i}_{y} \right)^T \mat{E} \mat{B} 
				+ \mat{A}^T \mat{E}^T \mat{E} \bv{W}{ls}{\bar{T}}{}{} \Lodot{}{\vec{i}_{y}} \LAtwo{W}{ls}{\epsilon}{} \\
			&= \left( \mat{E} \bv{W}{ls}{\bar{T}}{}{}  \vec{i}_{y} \right)^T \mat{E} \left(
					\bv{W}{p}{\bar{T}}{}{} \Lodot{}{\bv{p}{lh}{p}{}{}} \LAtwo{W}{p}{\epsilon}{}
					- \bv{W}{ls}{\bar{T}}{}{} \Lodot{}{\bv{ls}{lk}{p}{}{}} \LAtwo{W}{ls}{\epsilon}{}
				\right)
				+ \mat{A}^T \mat{E}^T \mat{E} \bv{W}{ls}{\bar{T}}{}{} \Lodot{}{\vec{i}_{y}} \LAtwo{W}{ls}{\epsilon}{} \\
		\mathcal{C}_{lkh,k} &= \left[ \begin{array}{c:c:c:c}
			\left( \mat{E} \bv{W}{ls}{\bar{T}}{}{}  \vec{i}_{y} \right)^T 		
				\mat{E} \bv{W}{p}{\bar{T}}{}{} \Lodot{}{\bv{p}{lh}{p}{}{}} & 
			-\left( \mat{E} \bv{W}{ls}{\bar{T}}{}{}  \vec{i}_{y} \right)^T 
				\mat{E} \bv{W}{ls}{\bar{T}}{}{} \Lodot{}{\bv{ls}{lk}{p}{}{}}
			+ \mat{A}^T \mat{E}^T \mat{E} \bv{W}{ls}{\bar{T}}{}{} \Lodot{}{\vec{i}_{y}} & 
			\mat{0}_{1 \times 6} & \mat{0}_{1 \times 9} \\
		\end{array} \right] \label{eq:C-lkhinge-k}
	\end{align}
	
\subsection{Knee range of motion}
	Thirdly, the constraint for the knee range of motion (ROM) is enforced if the knee angle is outside the allowed ROM.
		This implementation is similar to the active set method used in optimization.
		Mathematically, this is implemented by setting the constrained knee angle $\alpha_{lk}'$ to Eq. \eqref{eq:c-lkrom-cstr} where $\alpha_{lk,min}=0^\circ$ to prevent knee hyperextension and $\alpha_{lk,max}=\mathbf{min}(180^\circ, \hat{\alpha}^+_{lk})$.
		The knee angle $\alpha_{lk}$ is calculated by taking the inverse tangent of the thigh vector, $\bv{W}{lt}{r}{}{z}$, projected on the $z$ and $x$ axes of the shank orientation as shown in Eq. \eqref{eq:c-lkrom-base}.
		It ranges from $-\frac{\pi}{2}$ to $\frac{3\pi}{2}$ as enforced by the chosen sign inside the tangent inverse function and the addition of $\frac{\pi}{2}$.
	
	\begin{gather}
		\alpha_{lk}' = \mathbf{min}(\alpha_{lk,max}, \mathbf{max}(\alpha_{lk,min}, \alpha_{lk})) \label{eq:c-lkrom-cstr} \\
		\alpha_{lk} = \tan^{-1} \left( 
			\frac{ -\bv{W}{ls}{r}{T}{z} \bv{W}{lt}{r}{}{z} }{ -\bv{W}{ls}{r}{T}{x} \bv{W}{lt}{r}{}{z} } 
		\right) + \tfrac{\pi}{2} \label{eq:c-lkrom-base} \\
		\frac{ -\bv{W}{ls}{r}{T}{z} \bv{W}{lt}{r}{}{z} }{ -\bv{W}{ls}{r}{T}{x} \bv{W}{lt}{r}{}{z} } = 
			\frac{\sin(\alpha_{lk}' - \tfrac{\pi}{2})}{\cos(\alpha_{lk}' - \tfrac{\pi}{2})} \label{eq:c-lkrom-int1} \\
		\text{Let us now define this as our } c_{lkr} \text{ function} \nonumber \\
		\begin{split}
		c_{lkr}(\kfsc{\mu}{k}) &= 
		(\bv{W}{ls}{r}{T}{z} \tiny\cos(\alpha_{lk}' \text{--} \tfrac{\pi}{2}) \text{--} 
		\bv{W}{ls}{r}{T}{x} \tiny\sin(\alpha_{lk}' \text{--} \tfrac{\pi}{2}))
		\bv{W}{lt}{r}{}{z} \\
		&= ( \mat{E} \bv{W}{ls}{T}{}{} \overbrace{\left( 
			\vec{i}_z \cos(\alpha_{lk}' - \tfrac{\pi}{2}) 
			\vec{i}_x \sin(\alpha_{lk}' - \tfrac{\pi}{2}) \right)
		}^{\vec{\psi}} )^T \bv{W}{lt}{\tau}{}{z} \\
		&= 0 = \mat{D}_{lkr}
		\end{split} \label{eq:c-lkrom-int3}
		\end{gather}
		\begin{align}
 		\mat{\delta c}_{lkr} &= \Ltranvec{G_{lkr}}{ 
	 			c_{lkr} \left( \kfsm{\mu}{k} \right)^{-1} 
	 			c_{lkr} \left( \vec{\mu}_k^\epsilon \right) } \quad \text{ Note that the output is $\in \R^1$}  \\
 			&= c_{lkr} \left( \vec{\mu}_k^\epsilon \right) - c_{lkr} \left( \kfsp{\mu}{k} \right) \\
 			&= \left( \mat{E} \bv{W}{ls}{\bar{T}}{}{} \Lvectran{}{\LAtwo{W}{ls}{\epsilon}{}} \vec{\psi} \right)^T \mat{E} \left( \mat{A} + \mat{B} \right) 
	 			- \left( \mat{E} \bv{W}{ls}{\bar{T}}{}{} \vec{\psi} \right)^T \mat{E} \mat{A} \\
 			& \text{ Linearize } \exp(\vec{\phi}) \approx \mat{I} + \Lhat{}{\vec{\phi}} \nonumber \\
 			&= \left( \mat{E} \bv{W}{ls}{\bar{T}}{}{} \vec{\psi} \right)^T \mat{E} \left( \mat{A} + \mat{B} \right)
 				+ \left( \mat{E} \bv{W}{ls}{\bar{T}}{}{} \Lhat{}{\LAtwo{W}{ls}{\epsilon}{}} \vec{\psi} \right)^T \mat{E} \left( \mat{A} + \mat{B} \right)
	 			- \left( \mat{E} \bv{W}{ls}{\bar{T}}{}{} \vec{\psi} \right)^T \mat{E} \mat{A} \\
 			&= \left( \mat{E} \bv{W}{ls}{\bar{T}}{}{} \vec{\psi} \right)^T \mat{E} \mat{B} 
 				+ \left( \mat{E} \bv{W}{ls}{\bar{T}}{}{} \Lodot{}{\vec{\psi}} \LAtwo{W}{ls}{\epsilon}{} \right)^T \mat{E} \left( \mat{A} + \mat{B} \right) \\
 			& \text{Assume second order error } \approx 0 \nonumber \text{ and scalar so transposable} \\
			&= \left( \mat{E} \bv{W}{ls}{\bar{T}}{}{} \vec{\psi} \right)^T \mat{E} \mat{B} 
				+ \mat{A}^T \mat{E}^T \mat{E} \bv{W}{ls}{\bar{T}}{}{} \Lodot{}{\vec{\psi}} \LAtwo{W}{ls}{\epsilon}{} \\
			&= \left( \mat{E} \bv{W}{ls}{\bar{T}}{}{} \vec{\psi} \right)^T \mat{E} \left(
					\bv{W}{p}{\bar{T}}{}{} \Lodot{}{\bv{p}{lh}{p}{}{}} \LAtwo{W}{p}{\epsilon}{}
					- \bv{W}{ls}{\bar{T}}{}{} \Lodot{}{\bv{ls}{lk}{p}{}{}} \LAtwo{W}{ls}{\epsilon}{}
				\right)
				+ \mat{A}^T \mat{E}^T \mat{E} \bv{W}{ls}{\bar{T}}{}{} \Lodot{}{\vec{\psi}} \LAtwo{W}{ls}{\epsilon}{} \\
			\mathcal{C}_{lkr, k} &= \left[ \begin{array}{c:c:c:c}
				\left( \mat{E} \bv{W}{ls}{\bar{T}}{}{} \vec{\psi} \right)^T \mat{E} \bv{W}{p}{\bar{T}}{}{} \Lodot{}{\bv{p}{lh}{p}{}{}} & 
				- \left( \mat{E} \bv{W}{ls}{\bar{T}}{}{} \vec{\psi} \right)^T \mat{E} \bv{W}{ls}{\bar{T}}{}{} \Lodot{}{\bv{ls}{lk}{p}{}{}} 
					+ \mat{A}^T \mat{E}^T \mat{E} \bv{W}{ls}{\bar{T}}{}{} \Lodot{}{\vec{\psi}} & 
			\mat{0}_{1 \times 6} & \mat{0}_{1 \times 9} \\
			\end{array} \right] \label{eq:C-lkrom-k}
		\end{align}
		
	\printbibliography


\maketitle

\section{Additional details for Section II-C.1 Predication update} \label{sec:pred-update}
    Below is the explicit definition of the motion model $\Omega \left( \mat{X}{k} \right)$ and $\mathscr{C}_k$.
    
    \begin{gather}
        \begin{split}
        	\Omega ( \mat{X}_{k} ) = \big[
	        	(
	        	\dt \bv{W}{mp}{v}{}{k} +
	        	\tfrac{\dt^2}{2} \bvmeas{W}{p}{a}{}{k} 
	        	)^T \bvmeas{W}{p}{R}{}{k} \:\:
	        	\bv{}{}{0}{}{1 \times 3} \:\:
	        	(
	        	\dt \bv{W}{la}{v}{}{k} +
	        	\tfrac{\dt^2}{2} \bvmeas{W}{ls}{a}{}{k} 
	        	)^T \bvmeas{W}{ls}{R}{}{k} \:\:
	        	\bv{}{}{0}{}{1 \times 3} \\
	        	(
	        	\dt \bv{W}{ra}{v}{}{k} +
	        	\tfrac{\dt^2}{2} \bvmeas{W}{rs}{a}{}{k} 
	        	)^T \bvmeas{W}{rs}{R}{}{k} \:\:
	        	\bv{}{}{0}{}{1 \times 3} \:\:
	        	\dt \bvmeas{W}{mp}{a}{T}{k} \:\:
	        	\dt \bvmeas{W}{la}{a}{T}{k} \:\:
	        	\dt \bvmeas{W}{ra}{a}{T}{k}
        	\big]^T
       	\end{split} \\
       	\mathscr{C}_k 
       	    = \tfrac{\partial}{\partial \vec{\epsilon}} 
	    \Omega\left( \vec{\mu}_{k}^\epsilon \right) |_{\vec{\epsilon} = 0} 
	        = \left[ \begin{array}{c:ccc}
					\multirow{6}{*}{$\mat{0}_{18\times18}$} &
					\dt \bvmeas{W}{p}{R}{T}{k} & \quad\mat{0}_{3 \times 3}\quad & \quad\mat{0}_{3 \times 3} \quad \\
					& \mat{0}_{3 \times 3} & \mat{0}_{3 \times 3} & \mat{0}_{3 \times 3} \\
					& \mat{0}_{3 \times 3} & \dt \bvmeas{W}{ls}{R}{T}{k} & \mat{0}_{3 \times 3} \\
					& \mat{0}_{3 \times 3} & \mat{0}_{3 \times 3} & \mat{0}_{3 \times 3} \\
					& \mat{0}_{3 \times 3} & \mat{0}_{3 \times 3} & \dt \bvmeas{W}{rs}{R}{T}{k} \\
					& \mat{0}_{3 \times 3} & \mat{0}_{3 \times 3} & \mat{0}_{3 \times 3} \\ \hdashline
					\multicolumn{4}{c}{\raisebox{0pt}{$\mat{0}_{9 \times 27}$}}
				\end{array} \right]
    \end{gather}

\section{Additional details for Section II-C.2 Measurement update} \label{sec:meas-update}
    Only the derivation for $\mathcal{H}_{mp}$ will be shown below.
        The derivation for the other measurements are either trivial or can be solved similarly.
        The derivation for $\mathcal{H}_{ori}$ and $\mathcal{H}_{lim}$ are trivial as solving for $\LtranvecSmall{G_{a}}{
					h_{a}\left( \kfsp{\mu}{k} \right)^{-1} 
					h_{a}\left( \vec{\mu}_k^\epsilon \right) }$ where $a \in \{ ori, lim \}$
    		simply gives us the exponential coordinates of the corresponding perturbations, $\vec{\epsilon}$.
        The zero velocity part of $\mathcal{H}_{ls}$ and $\mathcal{H}_{rs}$ can also be calculated trivially, while the flat floor assumption can be calculated similarly as $\mathcal{H}_{mp}$ but the Z position set to floor height, $z_f$, instead of the pelvis standing height, $z_p$.
    
    Since the measurement function $h_{mp} ( \mat{X}_k ) \in \R$,  $\mat{X}_1^{-1} \mat{X}_2 = \mat{X}_2 - \mat{X}_1$.
        It then follows that $\mat{\delta h}_{mp} =
			\LtranvecSmall{G_{mp}}{ 
				h_{mp} ( \kfsp{\mu}{k} )^{-1} 
				h_{mp} ( \vec{\mu}_k^\epsilon ) } = h( \vec{\mu}_k^\epsilon ) - h( \kfsp{\mu}{k} )$;
			and that $\tfrac{\partial}{\partial \vec{\epsilon}} \mat{\delta h}_{mp} |_{\vec{\epsilon} = 0} = \tfrac{\partial}{\partial \vec{\epsilon}} h\left( \vec{\mu}_k^\epsilon \right) |_{\vec{\epsilon} = 0}$.
	    Also note of a useful property (Eq. \eqref{eq:se3-swap}) for $\vec{a} \in \LA{se(3)}, \vec{b} \in \R^4$ \cite[][Eq. (72)]{barfoot2017state}.
	
	\begin{gather}
		\Lhat{\LG{SE(3)}}{\bv{}{}{a}{}{}} \bv{}{}{b}{}{} = \bv{}{}{b}{}{} \Lodot{}{\bv{}{}{a}{}{}}, \label{eq:se3-swap} \quad
		\begin{bmatrix}
			\epsilon \\ \eta
		\end{bmatrix}^\odot = \begin{bmatrix}
			\eta \mat{I}_{3 \times 3} & -\Lhat{SO(3)}{\epsilon} \\
			\mat{0}_{1 \times 3} & \mat{0}_{1 \times 3}
		\end{bmatrix} \\
		\vec{i}_z = \begin{bmatrix} 0 & 0 & 1 & 0  \end{bmatrix}^T, \quad
		\vec{i}_0 = \begin{bmatrix} 0 & 0 & 0 & 1 \end{bmatrix}^T, \quad
		h_{mp} \left( \mat{X}_{k} \right) = \vec{i}_z^T \bv{W}{p}{T}{}{} \vec{i}_0
	\end{gather}
	\begin{align}
		h_{mp} ( \vec{\mu}_k^\epsilon ) &= \vec{i}_z^T \bv{W}{p}{\bar{T}}{}{} \Lvectran{SE(3)}{\LAtwobare{W}{p}{\epsilon}{}} \vec{i}_0 
			\qquad \text{ Linearize } \exp(\vec{\epsilon}) \approx \mat{I} + \Lhat{}{\vec{\epsilon}} \text{ where } \epsilon \approx 0 \text{ (very small).} \\
			&= \vec{i}_z^T \bv{W}{p}{\bar{T}}{}{} \Lhat{}{\LAtwobare{W}{p}{\epsilon}{}} \vec{i}_0 
			= \vec{i}_z^T \bv{W}{p}{\bar{T}}{}{} \Lodot{}{\vec{i}_0} \LAtwo{W}{p}{\epsilon}{} 
			\quad \text{ Use Eq. \eqref{eq:se3-swap} to swap $\epsilon$ to the right} \\
		\mathcal{H}_{mp} &= \tfrac{\partial}{\partial \vec{\epsilon}} h_{mp} ( \vec{\mu}_k^\epsilon ) |_{\vec{\epsilon} = 0}
		    = \left[ \begin{array}{ccc:c}
    			\vec{i}_z^T \bv{W}{p}{\bar{T}}{}{} \Lodot{}{\vec{i}_0} & \mat{0}_{1 \times 6} & \mat{0}_{1 \times 6} & \mat{0}_{1 \times 9} \\
    			\end{array} \right] \label{eq:H-mp-k}
	\end{align}
	
\section{Additional details for Section II-C.3 Constraint update} \label{sec:const-update}
\subsection{Thigh length}
	Below is the derivation of $\mathcal{C}_{ltl,k} = \tfrac{\partial}{\partial \vec{\epsilon}} c_{ltl} ( \vec{\mu}_k^\epsilon ) |_{\vec{\epsilon} = 0}$ obtained from the thigh length constraint (Eq. \eqref{eq:c-lthigh}) where $\bv{W}{lt}{\tau}{}{z}(\kfsc{\mu}{k})$ is the thigh vector (Eq. \eqref{eq:thigh-vect}).
		$\mathcal{C}_{rtl,k}$ is derived similarly.
	\begin{gather}
		\mat{E} = \begin{bmatrix}
				\mat{I}_{3 \times 3} & \mat{0}_{3 \times 1}
			\end{bmatrix} \quad
		\bv{p}{lh}{p}{}{} = \begin{bmatrix}
				0 & \tfrac{d^{\cs{p}}}{2} & 0 & 1
			\end{bmatrix}^T \quad
		\bv{ls}{lk}{p}{}{} = \begin{bmatrix}
				0 & 0 & d^{\cs{ls}} & 1
			\end{bmatrix}^T \\
		\bv{W}{lt}{\tau}{}{z}(\kfsc{\mu}{k}) = 
			\overbrace{\mat{E} \bv{W}{p}{T}{}{} \bv{p}{lh}{p}{}{} }^{\text{hip joint pos.}} - 
			\overbrace{\mat{E} \bv{W}{ls}{T}{}{} \bv{ls}{lk}{p}{}{} }^{\text{knee joint pos.}} \label{eq:thigh-vect} \\
		c_{ltl}(\kfsc{\mu}{k}) = \bv{W}{lt}{\tau}{}{z}(\kfsc{\mu}{k})^T 			
			\bv{W}{lt}{\tau}{}{z}(\kfsc{\mu}{k}) - (d^{\cs{lt}})^2 = 0 = \mat{D}_{ltl} \label{eq:c-lthigh}
	\end{gather}
	\begin{align}
            & \text{For simplicity let us first define } \bv{W}{lt}{\tau}{}{z} ( \vec{\mu}_k^\epsilon ) \text{ and linearize } \exp(\vec{\epsilon}) \approx \mat{I} + \Lhat{}{\vec{\epsilon}} \\
		\bv{W}{lt}{\tau}{}{z} ( \vec{\mu}_k^\epsilon ) & = \mat{E} ( 
				\bv{W}{p}{\bar{T}}{}{} \LvectranSmall{}{\LAtwo{W}{p}{\epsilon}{}} \bv{p}{lh}{p}{}{} - \bv{W}{ls}{\bar{T}}{}{} \LvectranSmall{}{\LAtwo{W}{ls}{\epsilon}{}} \bv{ls}{lk}{p}{}{}
			) \\
			& = \mat{E} ( 
				\bv{W}{p}{\bar{T}}{}{} \bv{p}{lh}{p}{}{} 
				- \bv{W}{ls}{\bar{T}}{}{} \bv{ls}{lk}{p}{}{}
				+ \bv{W}{p}{\bar{T}}{}{} \Lhat{}{\LAtwo{W}{p}{\epsilon}{}} \bv{p}{lh}{p}{}{}
				- \bv{W}{ls}{\bar{T}}{}{} \Lhat{}{\LAtwo{W}{ls}{\epsilon}{}} \bv{ls}{lk}{p}{}{}
			) \\
			& = \mat{E} \big( 
			\overbrace{
					\bv{W}{p}{\bar{T}}{}{} \bv{p}{lh}{p}{}{} 
					- \bv{W}{ls}{\bar{T}}{}{} \bv{ls}{lk}{p}{}{}
			}^{\mat{A}}
			+ \overbrace{
				\bv{W}{p}{\bar{T}}{}{} \Lodot{}{\bv{p}{lh}{p}{}{}} \LAtwo{W}{p}{\epsilon}{}
				- \bv{W}{ls}{\bar{T}}{}{} \Lodot{}{\bv{ls}{lk}{p}{}{}} \LAtwo{W}{ls}{\epsilon}{}
			}^{\mat{B}}
			\big) \\
			& \text{Calculating for }  c_{ltl} ( \vec{\mu}_k^\epsilon ) \text{ and noting that } \mat{A}^T \mat{E}^T \mat{E} \mat{B} = \mat{B}^T \mat{E}^T \mat{E} \mat{A} \text{ since it is scalar} \nonumber \\
		 c_{ltl} ( \vec{\mu}_k^\epsilon ) &= \left( \mat{A} + \mat{B} \right)^T \mat{E}^T \mat{E} \left( \mat{A} + \mat{B} \right) 
			- (d^{\cs{lt}})^2 \\
			&= \mat{A}^T \mat{E}^T \mat{E} \mat{A} + 2 \mat{A}^T \mat{E}^T \mat{E} \mat{B} + \mat{B}^T \mat{E}^T \mat{E} \mat{B} - (d^{\cs{lt}})^2 \\
			& \text{Assume second order error } \mat{B}^T \mat{E}^T \mat{E} \mat{B} \approx 0 \nonumber \\
			&= \mat{A}^T \mat{E}^T \mat{E} \mat{A} + 2 \mat{A}^T \mat{E}^T \mat{E} \big(
				\bv{W}{p}{\bar{T}}{}{} \Lodot{}{\bv{p}{lh}{p}{}{}} \LAtwo{W}{p}{\epsilon}{}
				- \bv{W}{ls}{\bar{T}}{}{} \Lodot{}{\bv{ls}{lk}{p}{}{}} \LAtwo{W}{ls}{\epsilon}{}
			\big) - (d^{\cs{lt}})^2\\
		\mathcal{C}_{ltl,k} &= \tfrac{\partial}{\partial \vec{\epsilon}} c_{ltl} ( \vec{\mu}_k^\epsilon ) |_{\vec{\epsilon} = 0}
		    = \left[ \begin{array}{ccc:c}
    			2 \mat{A}^T \mat{E}^T \mat{E} \bv{W}{p}{\bar{T}}{}{} \Lodot{}{\bv{p}{lh}{p}{}{}} & 
    			-2 \mat{A}^T \mat{E}^T \mat{E} \bv{W}{ls}{\bar{T}}{}{} \Lodot{}{\bv{ls}{lk}{p}{}{}} & 
    			\mat{0}_{1 \times 6} & \mat{0}_{1 \times 9} \\
    		\end{array} \right] \label{eq:C-ltl-k}
	\end{align}
	
\subsection{Hinge knee joint}
    Below is the derivation of $\mathcal{C}_{lkh,k} = \tfrac{\partial}{\partial \vec{\epsilon}} c_{lkh} ( \vec{\mu}_k^\epsilon ) |_{\vec{\epsilon} = 0}$ obtained from the constraint for the hinge knee joint (Eq. \eqref{eq:c-lkhinge}).
		$\mathcal{C}_{rkh,k}$ is derived similarly.

	\begin{gather}
	    \vec{i}_y = \begin{bmatrix} 0 & 1 & 0 & 0  \end{bmatrix}^T, \quad
		c_{lkh}(\vec{\mu}_k) = \bv{W}{ls}{r}{T}{y} \bv{W}{lt}{\tau}{}{z} \label{eq:c-lkhinge} 
		    = \left( \mat{E} \bv{W}{ls}{T}{}{} \vec{i}_{y} \right)^T \bv{W}{lt}{\tau}{}{z}
			= 0 = \mat{D}_{lkh}
	\end{gather}
	\begin{align}
	        & \text{ Linearize } \exp(\vec{\epsilon}) \approx \mat{I} + \Lhat{}{\vec{\epsilon}} \nonumber \\
		c_{lkh} \left( \vec{\mu}_k^\epsilon \right)
		    &= ( \mat{E} \bv{W}{ls}{\bar{T}}{}{} \LvectranSmall{}{\LAtwo{W}{ls}{\epsilon}{}} \vec{i}_{y} )^T \mat{E} ( \mat{A} + \mat{B} ) 
			= ( \mat{E} ( \bv{W}{ls}{\bar{T}}{}{}  + \bv{W}{ls}{\bar{T}}{}{} \Lhat{}{\LAtwo{W}{ls}{\epsilon}{}} ) \vec{i}_{y} )^T \mat{E} \left( \mat{A} + \mat{B} \right) \\
			&= \left( \mat{E} \bv{W}{ls}{\bar{T}}{}{} \vec{i}_{y} \right)^T \mat{E} \left(\mat{A} + \mat{B}\right) 
			+ ( \mat{E} \bv{W}{ls}{\bar{T}}{}{} \Lhat{}{\LAtwo{W}{ls}{\epsilon}{}} \vec{i}_{y} )^T \mat{E} \left(\mat{A} + \mat{B}\right) \\
			& \text{Assume second order error } \approx 0  \text{, scalar so transposable, and using Eq. \eqref{eq:se3-swap}} \nonumber \\
			&= \left( \mat{E} \bv{W}{ls}{\bar{T}}{}{}  \vec{i}_{y} \right)^T \mat{E} \left(\mat{A} + \mat{B}\right) 
				+ \mat{A}^T \mat{E}^T \mat{E} \bv{W}{ls}{\bar{T}}{}{} \Lodot{}{\vec{i}_{y}} \LAtwo{W}{ls}{\epsilon}{} \\
			&= \left( \mat{E} \bv{W}{ls}{\bar{T}}{}{}  \vec{i}_{y} \right)^T \mat{E} (
			        \mat{A} + 
					\bv{W}{p}{\bar{T}}{}{} \Lodot{}{\bv{p}{lh}{p}{}{}} \LAtwo{W}{p}{\epsilon}{}
					- \bv{W}{ls}{\bar{T}}{}{} \Lodot{}{\bv{ls}{lk}{p}{}{}} \LAtwo{W}{ls}{\epsilon}{}
				)
				+ \mat{A}^T \mat{E}^T \mat{E} \bv{W}{ls}{\bar{T}}{}{} \Lodot{}{\vec{i}_{y}} \LAtwo{W}{ls}{\epsilon}{} \\
		\mathcal{C}_{lkh,k} &= \left[ \begin{array}{c:c:c:c}
			\left( \mat{E} \bv{W}{ls}{\bar{T}}{}{}  \vec{i}_{y} \right)^T 		
				\mat{E} \bv{W}{p}{\bar{T}}{}{} \Lodot{}{\bv{p}{lh}{p}{}{}} & 
			\text{--}\left( \mat{E} \bv{W}{ls}{\bar{T}}{}{}  \vec{i}_{y} \right)^T 
				\mat{E} \bv{W}{ls}{\bar{T}}{}{} \Lodot{}{\bv{ls}{lk}{p}{}{}}
			+ \mat{A}^T \mat{E}^T \mat{E} \bv{W}{ls}{\bar{T}}{}{} \Lodot{}{\vec{i}_{y}} & 
			\mat{0}_{1 \times 6} & \mat{0}_{1 \times 9} \\
		\end{array} \right] \label{eq:C-lkhinge-k}
	\end{align}
	
\subsection{Knee range of motion}
    Below is the derivation of $\mathcal{C}_{lkr,k} = \tfrac{\partial}{\partial \vec{\epsilon}} c_{lkr} ( \vec{\mu}_k^\epsilon ) |_{\vec{\epsilon} = 0}$ obtained from the constraint for the knee range of motion (ROM) which is enforced if the knee angle is outside the allowed ROM (Eq. \eqref{eq:c-lkrom-int3}).
		$\mathcal{C}_{rkr,k}$ is derived similarly.
		
	\begin{gather}
		\begin{split}
		c_{lkr}(\kfsc{\mu}{k}) &= 
		(\bv{W}{ls}{r}{T}{z} \tiny\cos(\alpha_{lk}' - \tfrac{\pi}{2}) - 
		\bv{W}{ls}{r}{T}{x} \tiny\sin(\alpha_{lk}' - \tfrac{\pi}{2}))
		\bv{W}{lt}{r}{}{z} \\
		&= ( \mat{E} \bv{W}{ls}{T}{}{} \overbrace{\left( 
			\vec{i}_z \cos(\alpha_{lk}' - \tfrac{\pi}{2}) 
			\vec{i}_x \sin(\alpha_{lk}' - \tfrac{\pi}{2}) \right)
		}^{\vec{\psi}} )^T \bv{W}{lt}{\tau}{}{z} = 0 = \mat{D}_{lkr}
		\end{split} \label{eq:c-lkrom-int3}
		\end{gather}
		\begin{align}
            c_{lkr} \left( \vec{\mu}_k^\epsilon \right)
 			&= ( \mat{E} \bv{W}{ls}{\bar{T}}{}{} \LvectranSmall{}{\LAtwo{W}{ls}{\epsilon}{}} \vec{\psi} )^T \mat{E} \left( \mat{A} + \mat{B} \right)
 			\quad \text{ Linearize } \exp(\vec{\epsilon}) \approx \mat{I} + \Lhat{}{\vec{\epsilon}} \nonumber \\
 			&= \left( \mat{E} \bv{W}{ls}{\bar{T}}{}{} \vec{\psi} \right)^T \mat{E} \left( \mat{A} + \mat{B} \right)
 				+ ( \mat{E} \bv{W}{ls}{\bar{T}}{}{} \Lhat{}{\LAtwo{W}{ls}{\epsilon}{}} \vec{\psi} )^T \mat{E} \left( \mat{A} + \mat{B} \right) \\
 			& \text{Assume second order error } \approx 0  \text{, scalar so transposable, and using Eq. \eqref{eq:se3-swap}} \nonumber \\
			&= \left( \mat{E} \bv{W}{ls}{\bar{T}}{}{} \vec{\psi} \right)^T \mat{E} \left( \mat{A} + \mat{B} \right) 
				+ \mat{A}^T \mat{E}^T \mat{E} \bv{W}{ls}{\bar{T}}{}{} \Lodot{}{\vec{\psi}} \LAtwo{W}{ls}{\epsilon}{} \\
			&= \left( \mat{E} \bv{W}{ls}{\bar{T}}{}{} \vec{\psi} \right)^T \mat{E} (
			        \mat{A} +
					\bv{W}{p}{\bar{T}}{}{} \Lodot{}{\bv{p}{lh}{p}{}{}} \LAtwo{W}{p}{\epsilon}{}
					- \bv{W}{ls}{\bar{T}}{}{} \Lodot{}{\bv{ls}{lk}{p}{}{}} \LAtwo{W}{ls}{\epsilon}{}
				)
				+ \mat{A}^T \mat{E}^T \mat{E} \bv{W}{ls}{\bar{T}}{}{} \Lodot{}{\vec{\psi}} \LAtwo{W}{ls}{\epsilon}{} \\
			\mathcal{C}_{lkr, k} &= \left[ \begin{array}{c:c:c:c}
				\left( \mat{E} \bv{W}{ls}{\bar{T}}{}{} \vec{\psi} \right)^T \mat{E} \bv{W}{p}{\bar{T}}{}{} \Lodot{}{\bv{p}{lh}{p}{}{}} & 
				\text{--} \left( \mat{E} \bv{W}{ls}{\bar{T}}{}{} \vec{\psi} \right)^T \mat{E} \bv{W}{ls}{\bar{T}}{}{} \Lodot{}{\bv{ls}{lk}{p}{}{}} 
					+ \mat{A}^T \mat{E}^T \mat{E} \bv{W}{ls}{\bar{T}}{}{} \Lodot{}{\vec{\psi}} & 
			\mat{0}_{1 \times 6} & \mat{0}_{1 \times 9} \\
			\end{array} \right] \label{eq:C-lkrom-k}
		\end{align}
		
	\printbibliography